\PassOptionsToPackage{table}{xcolor}
\documentclass{fairmeta}

\usepackage{amsmath,amsfonts,bm}

\def\eqref#1{equation~\ref{#1}}

\def\1{\bm{1}}

\DeclareMathAlphabet{\mathsfit}{\encodingdefault}{\sfdefault}{m}{sl}
\SetMathAlphabet{\mathsfit}{bold}{\encodingdefault}{\sfdefault}{bx}{n}

\usepackage{url}
\usepackage{amsmath,amssymb}
\usepackage{wrapfig}
\usepackage{listings}
\usepackage{tabularx}
\usepackage{makecell}
\usepackage{colortbl}
\usepackage{pifont}
\usepackage[ruled,linesnumbered]{algorithm2e}
\setcitestyle{round,authoryear}

\newtcblisting{promptbox}[1]{%
  listing only, enhanced, breakable,
  colback=gray!4, colframe=gray!60, boxrule=0.5pt, arc=3mm,
  left=5pt, right=5pt, top=4pt, bottom=4pt,
  fonttitle=\bfseries\footnotesize, coltitle=black, colbacktitle=gray!18,
  title={#1},
  listing options={basicstyle=\ttfamily\scriptsize, columns=fullflexible,
    breaklines=true, breakindent=0pt, breakautoindent=false, keepspaces=true,
    showstringspaces=false, frame=none, backgroundcolor=\color{gray!4},
    aboveskip=0pt, belowskip=0pt}}

\newcommand{\cmark}{\textcolor{green!55!black}{\ding{51}}}       
\newcommand{\xmark}{\textcolor{red!70!black}{\ding{55}}}         
\newcommand{\pmark}{\textcolor{orange!90!black}{\ding{119}}}     
\newcommand{\firstlogo}{\raisebox{-1ex}{\includegraphics[height=3.5ex]{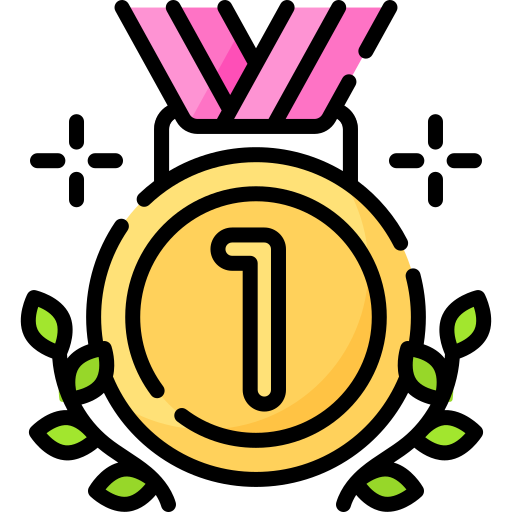}}}
\newcommand{\secondlogo}{\raisebox{-1ex}{\includegraphics[height=3.5ex]{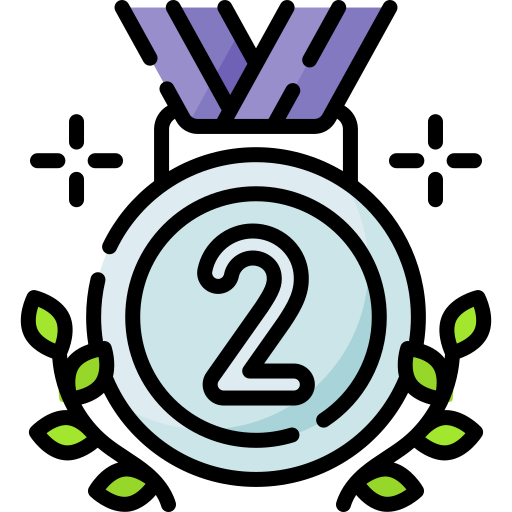}}}
\newcommand{\thirdlogo}{\raisebox{-1ex}{\includegraphics[height=3.5ex]{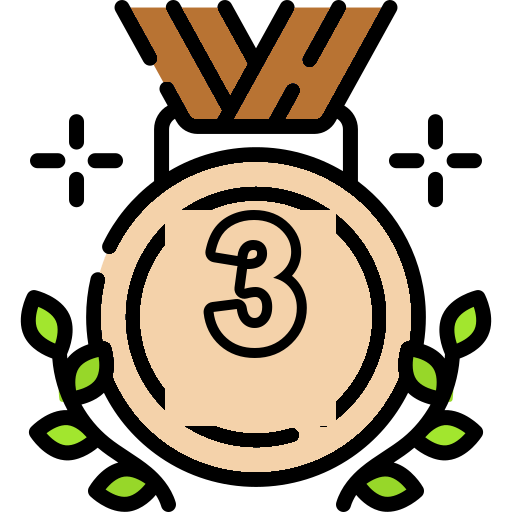}}}

\graphicspath{{Figures/}}

\usepackage{adjustbox}

\headerleft{\includegraphics[height=15pt]{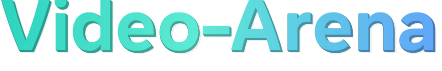}}
\headerright{\includegraphics[height=22pt]{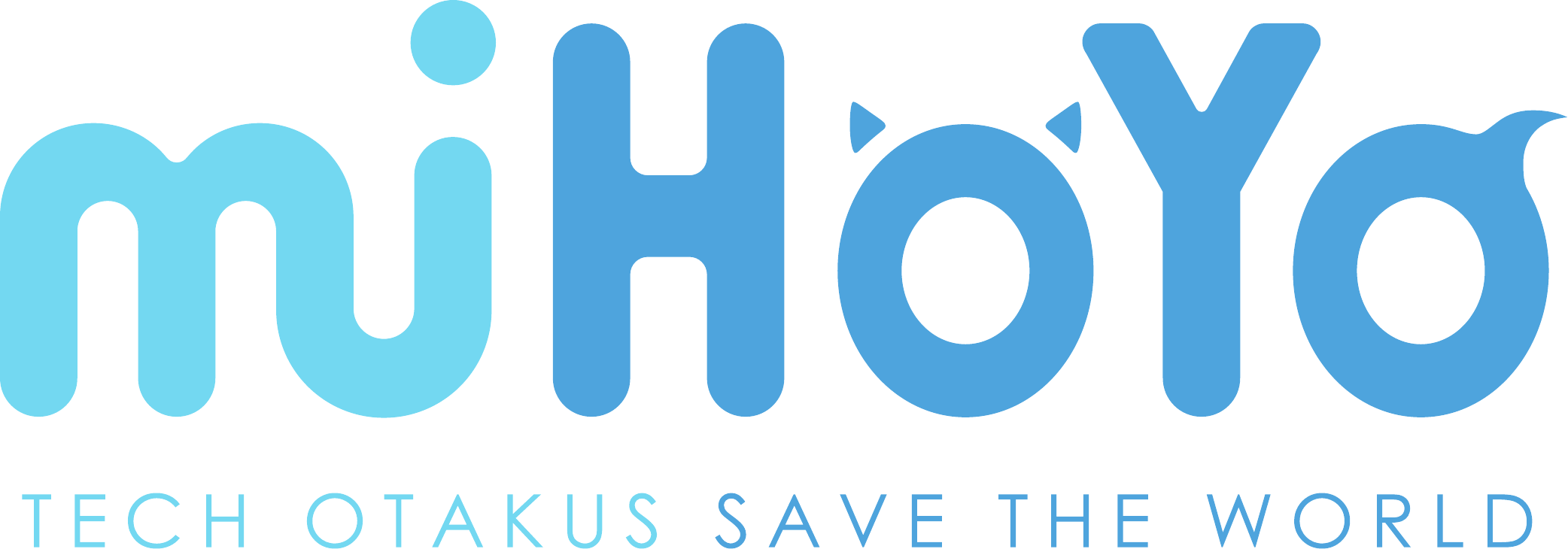}\hspace{8pt}\includegraphics[height=20pt]{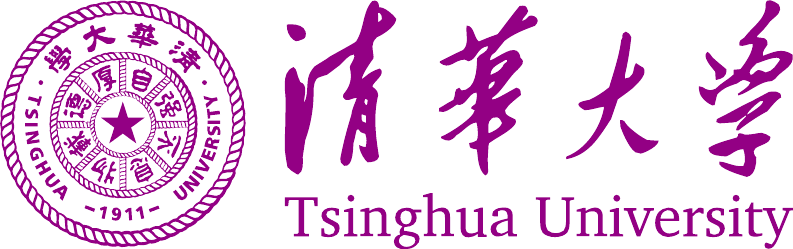}\hspace{8pt}\includegraphics[height=20pt]{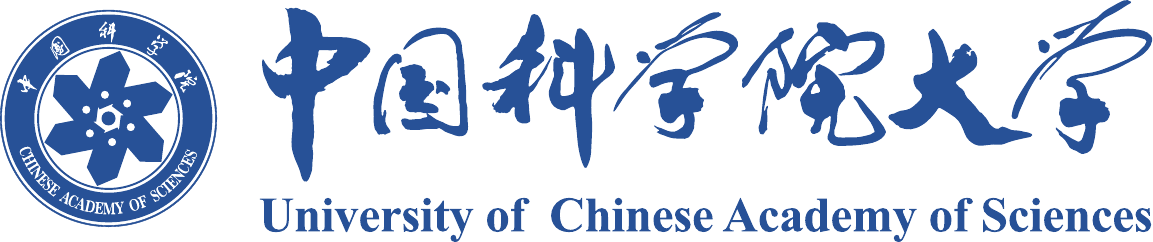}}

\title{\Large Video-Arena: A Scalable Agentic Video Benchmark for Spatiotemporal Capability}

\author[1,2,*,\ddagger]{Kuan Zhang}
\author[1,3,*,\ddagger]{Yukun Chen}
\author[1,3,*,\ddagger]{Zhihao Yang}
\author[4]{Yue Su}
\author[3]{Xiangnan Wu}
\author[2]{Zirong Chen}
\author[1,5,\ddagger]{Run Luo}
\author[1]{Tao Tan}
\author[1,\dagger]{Yinhe Zheng}
\author[2,\dagger]{Yiming Li}
\renewcommand\affiliationformat[2][]{{\small $^{#1}$#2}}
\affiliation[1]{miHoYo Honkai AI R\&D Team}
\affiliation[2]{College AI, Tsinghua University}
\affiliation[3]{University of Chinese Academy of Sciences}
\affiliation[4]{MMLab, The University of Hong Kong}
\affiliation[5]{National University of Singapore}
\contribution[*]{Equal Contribution}
\contribution[\dagger]{Corresponding Author}
\contribution[\ddagger]{Work done while interning at miHoYo}

\abstract{Understanding and tracking a dynamic world are fundamental to visual intelligence. Existing video benchmarks provide limited insight into whether models can maintain a coherent understanding of evolving world state. Building such a benchmark from real-world videos is challenging because the underlying world state is partially observable, making annotations hard to obtain. To close this gap, we introduce \textbf{Video-Arena}, a benchmark that turns competitive games into scalable, verifiable tests of how well models understand evolving world states from videos. Specifically, inspired by how humans understand dynamic events, we evaluate spatiotemporal world state understanding and tracking through three task categories: perceiving critical events, interpreting state updates, and tracking the resulting states. Our pipeline can generate videos and aligned reference labels from game records \emph{at scale}. To keep evaluation affordable and reproducible, we curate a golden set of 180 tasks across 9 games and run evaluations under two protocols: native visual understanding and tool-assisted agentic analysis. The best-performing systems achieve F1 scores of $13.2$ and $51.0$ under the two protocols, respectively. The results highlight that native models struggle to perceive events reliably, while tool-assisted systems improve event recovery but still struggle to interpret state updates and track the resulting states consistently over time.
\vspace{-6pt}}

\metadata[Project Page]{\url{https://videoarena-bench.github.io}}
\metadata[Correspondence]{\email{yimingli9702@mail.tsinghua.edu.cn}, \email{yinhe.zheng@mihoyo.com}}

\begin{document}

\maketitle

\section{Introduction}
\label{sec:intro}

\begin{figure}[!htbp]
\centering
\includegraphics[width=0.85\linewidth]{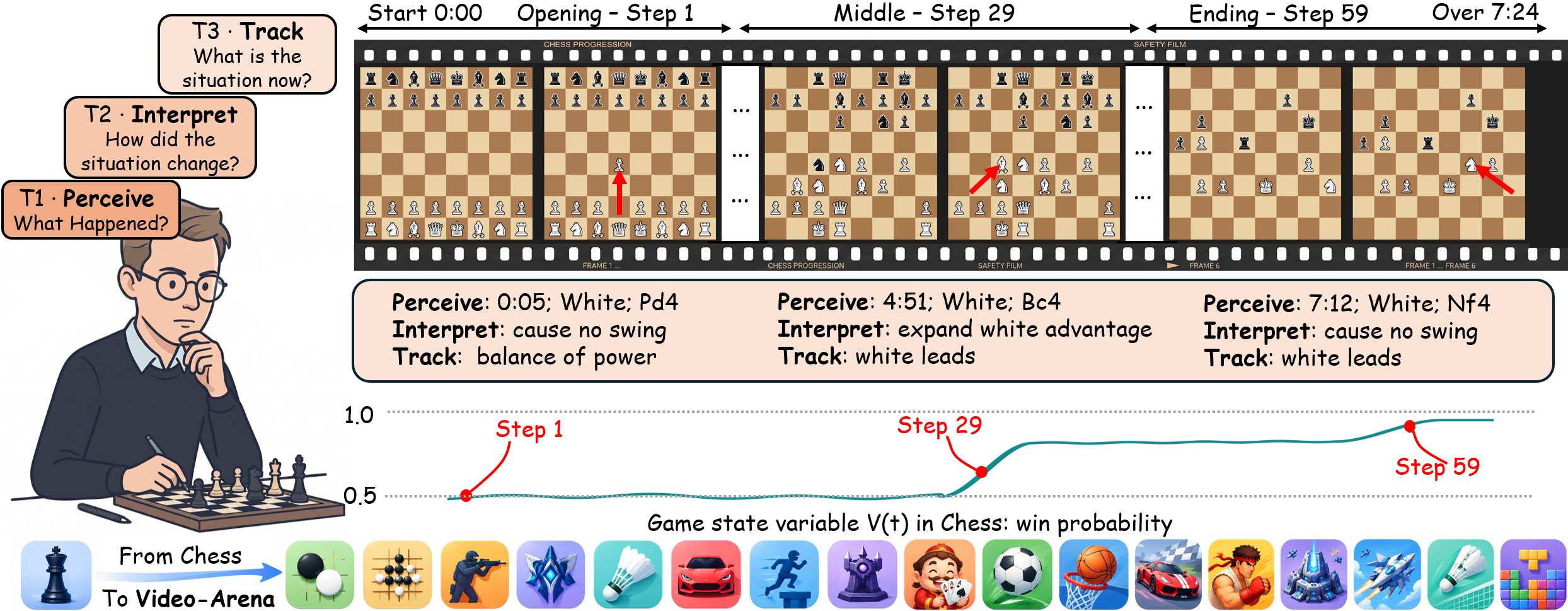}
\caption{Understanding an evolving chess game: T1, \emph{Perceiving} moves and captures; T2, \emph{Interpreting} which side benefit from each move; T3, \emph{Tracking} which side leads after each move. The tasks connect event perception, state interpretation, and tracking in a unified evaluation framework.}
\label{fig:human-strategist}
\end{figure}

Understanding how a dynamic world evolves is fundamental to visual intelligence. Cognitive research offers a foundation for studying this ability: humans organize continuous experience into events, integrate new evidence to update their belief, and maintain these representations over time \citep{zwaan1999situation,zacks2007event,hasson2008hierarchy}. We define this ability as \textbf{\textit{spatiotemporal world state understanding and tracking}}, essential for multimodal agents in dynamic environments.

Recent multimodal models have achieved strong performance on video benchmarks \citep{videomme, longvideobench}. However, existing benchmarks rarely quantify the ability to understand the evolving world.  
Manual questions may be answerable from a single frame, subtitle or event without visual inputs~\citep{mmstar, vsisuperwild, mirage}; multiple-choice questions also expose priors, noise and contamination~\citep{dataconbination, noisylabel}. Therefore, the question is not video duration itself, but whether a benchmark forces a model to track an evolving world state correctly. Constructing such a benchmark from real-world videos is challenging: underlying states may be  partially observable, and aligned symbolic states are difficult to derive from visual evidence alone, even with the help of expensive manual annotations \citep{chen2026snapbench}. Table~\ref{tab:landscape} positions Video-Arena against representative video, gameplay, and strategy-video benchmarks along these axes: none of the existing entries is simultaneously video-only, gameplay-grounded, resistant to answer leakage, and equipped with automated ground truth for tracking an evolving world state.

\begin{table}[t]
\centering
\footnotesize
\setlength{\tabcolsep}{4.5pt}
\renewcommand{\arraystretch}{1.15}
\caption{Comparison with related benchmarks. \emph{Video-only}: video input without symbolic logs or standalone images. \emph{No-Action}: observation without gameplay. \emph{Anti-shortcut}: controls against answer leakage and contamination. \emph{Auto.GT}: automated ground truth without human or LLM judge. \emph{World state}: tracking evolving states. (\cmark/\pmark/\xmark~= yes/partial/no) }
\label{tab:landscape}
\adjustbox{max width=0.83\linewidth}{%
\begin{tabular}{@{}cl ccccc c@{}}
\toprule
\textbf{Family} & \textbf{Benchmark} & Video-only & No-Action & Anti-shortcut & Auto.GT & World State & Len (min) \\
\midrule
\multirow{3}{*}{\makecell{General\\Video\\Understanding}}
 & LongVideoBench~\citep{longvideobench}& \cmark & \cmark & \xmark & \xmark & \xmark & 0.1$\sim$60 \\
 & Video-MME~\citep{videomme}          & \cmark & \cmark & \xmark & \xmark & \xmark & 0.2$\sim$60 \\
 & VSI-Super~\citep{cambrians}          & \cmark & \cmark & \xmark & \cmark & \pmark & 10$\sim$240 \\
 
\midrule
\multirow{3}{*}{\makecell{Video\\Game\\Play}}
 & GameBench~\citep{gamebench}          & \pmark & \xmark & \cmark & \cmark & \pmark & --- \\
 & GameVerse~\citep{gameverse}          & \pmark & \xmark & \cmark & \cmark & \pmark & --- \\
 & GameWorld~\citep{gameworld}          & \pmark & \xmark & \cmark & \cmark & \pmark & --- \\
\midrule
\multirow{6}{*}{\makecell{Strategy\\Video\\Intelligence}}
 & LoL-V2T~\citep{lolv2t}               & \cmark & \cmark & \xmark & \xmark & \xmark & 0.1$\sim$4.6 \\
 & Game-MUG~\citep{gamemug}             & \xmark & \cmark & \xmark & \pmark & \xmark & --- \\
 & EgoEsportsQA~\citep{egoesportsqa}    & \cmark & \cmark & \pmark & \xmark & \xmark & $\leq$3.3 \\
 & SVI-Bench~\citep{svibench}           & \pmark & \cmark & \xmark & \pmark & \pmark & 0.2$\sim$180 \\
 & PRS-Diag~\citep{seeingwithout}& \xmark & \cmark & \pmark & \cmark & \xmark & --- \\
 \midrule
\rowcolor{blue!8}
 & \textbf{Video-Arena (ours)}         & \cmark & \cmark & \cmark & \cmark & \cmark & 1.3$\sim$88 \\
\bottomrule
\end{tabular}%
}
\end{table}

Fortunately, games offer a controlled environment for this challenge \citep{gamebench, svibench, gamesurvey}. For each time step $i$, we can easily recover game-related state variable $V_i$ from replay records and evaluators, which are readily available and can be automatically scaled.
Tracking the evolution of $V_i$ yields a coupled capability: a model must first \emph{perceive} what happened and when; then \emph{interpret} the associated change in $V_i$; and finally \emph{track} the resulting state over time. 
Figure~\ref{fig:human-strategist} illustrates this process in chess.
They make spatiotemporal world state understanding and tracking measurable by testing whether a model can recover how a game state evolves. 

In this paper, we introduce \textbf{Video-Arena}, a benchmark for reconstructing spatiotemporal world states from game videos. Specifically, we design three kinds of coupled tasks based on the spatiotemporal capability: \textit{\textbf{T1 (Perceive)}} critical events; \textbf{\textit{T2 (Interpret)}} the associated changes in $V_i$; and \textit{\textbf{T3 (Track)}} the resulting states over time. Our pipeline converts game records into videos and derives aligned reference labels for each task using a deterministic verifier based on game engines. This automatic pipeline benefits from a large and growing supply of game records, enabling scalable evaluation without additional manual annotation or LLM judges. To keep evaluation affordable and reproducible, we curate a representative \emph{golden set} of 180 tasks across 9 games, with 20 tasks per game and an average duration of 24.3 minutes. We evaluate models under native visual and tool-assisted agentic protocols. Results show that frontier models struggle to perceive critical events. Agentic tools improve event perception, but models still struggle to interpret state updates and track the resulting states over time. Our benchmark Video-Arena reveals a substantial performance gap between frontier models and reliable spatiotemporal world-state understanding and tracking.

Our major contributions are summarized as follows:
\begin{itemize}
  \item \textbf{A cognitively motivated evaluation framework.} We turn spatiotemporal world state understanding and tracking into a measurable test with three kinds of coupled tasks: perceiving critical events, interpreting state updates, and tracking the resulting states.
  
  \item \textbf{A scalable benchmark with a representative golden set.} We develop an automated pipeline that generates videos and reference labels by records and evaluators. This pipeline benefits from a large and continually growing game supply. To keep evaluations affordable and reproducible, we curate a golden set of 180 tasks across 9 complementary games.
  
\item \textbf{Evidence of a persistent performance gap.} We show that frontier models struggle to perceive critical events reliably. Tool assistance improves event recovery, but interpreting state updates and tracking the resulting states remain challenging.

\end{itemize}

\section{Related Work}
\label{sec:related}
\begin{figure}[t]
\centering
\includegraphics[width=0.87\linewidth]{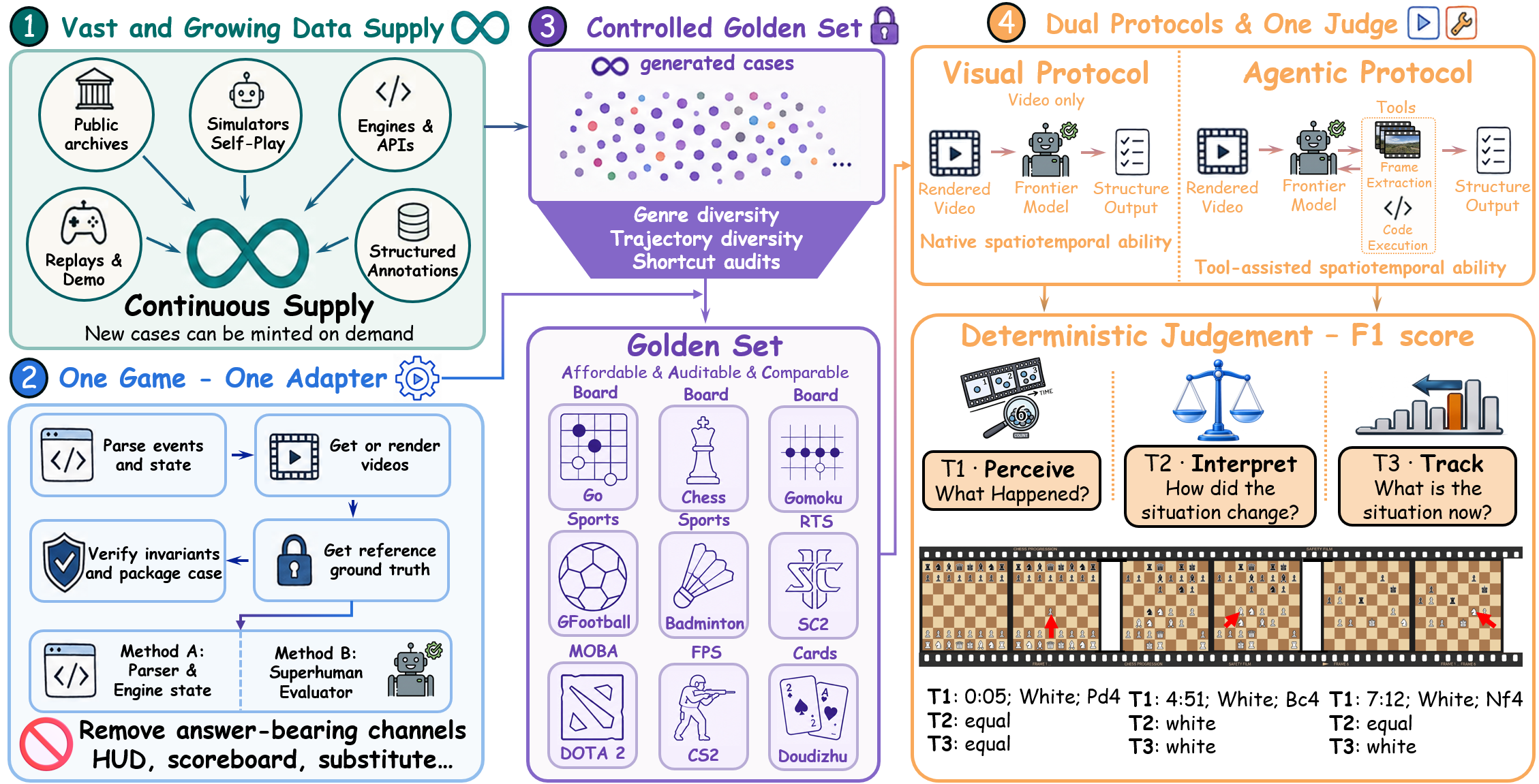}
\caption{\textbf{Overview of Video-Arena.} A vast supply of game records support automatic benchmark construction without additional per-case annotation, requiring only one adapter per new game. We curated a controlled golden set to keep evaluation affordable and comparable. Two evaluation protocols use a deterministic verifier, eliminating the need for human and LLM judges.}
\label{fig:pipeline}
\end{figure}

\textbf{Multimodal foundation models and agents.}
Foundation models have made rapid progress in video understanding \citep{gpt4o, gemini2.5, qwen3vl}. However, their success on conventional video tasks may rely on static visual cues or language priors, leaving substantial gaps in temporal reasoning \citep{tvbench,tempcompass}. More advanced visual intelligence requires interpreting the significance of critical events and integrating observations over time \citep{model1, model2, model3}. Whether frontier vision systems can reliably perceive events and maintain a coherent understanding of an evolving world remains an open question. 

\textbf{Video understanding benchmarks.} Video benchmarks are essential for assessing model capabilities. Recent works extend to longer videos and broader range of tasks \citep{videomme,longvideobench,mlvu}. However, issues such as single-frame bias, reliance on language priors, and questions answerable without visual input still persist \citep{mmstar, mirage}, limiting existing benchmarks to reliably assess spatiotemporal world-state understanding and tracking \citep{cambrians, vsisuperwild}. Moreover,  high quality annotations for long videos require substantial human effort in question writing, and verification \citep{svibench, seeingwithout}.

\textbf{Sports and game understanding.}
Several studies use sport or game environments to evaluate video understanding~\citep{lolv2t, gamemug, proact-vl}, providing structured event logs instead of game state recovery. There are also game-playing benchmarks that evaluate agents in interactive games~\citep{gamebench, gameworld, gameverse}, assessing control and latency in addition to understanding. Three recent works are related to ours: SVI-Bench \citep{svibench} reveals the leading side via textual information rather than visual reasoning; EgoEsportsQA \citep{egoesportsqa} relies on multiple-choice items over short clips; and PRS-Diag \citep{seeingwithout} evaluates board states without temporal tracking. While these works examine different aspects of game understanding, they do not jointly evaluate event perception, state interpretation and state tracking from video. Video-Arena brings these capabilities together, requiring models to connect what happened with how the game state changed and which state resulted.

\section{The Video-Arena Benchmark}
\label{sec:bench}
\begin{table}[th!]
\centering
\caption{The autonomous Video-Arena data pipeline. \emph{Supply}: publicly available records, $\dagger$ marks an estimate. \emph{State label generator}: \textbf{parser} = events and state recorded by the engine; \textbf{evaluator} = a superhuman engine scores the state. Detailed setup and costs are in Appendix~\ref{app:envs}.}
\label{tab:flywheel}
\adjustbox{max width=\linewidth}{%
\begin{tabular}{llll}
\toprule
\textbf{Game} & \textbf{Record} & \textbf{Supply} & \textbf{State label generator} \\
\midrule
\rowcolor{gray!15}
\multicolumn{4}{l}{\emph{Golden Set: composed of 9 popular and comprehensive game environments.}}\\
Go            & board record (SGF), re-rendered              & $10^{7\,\dagger}$      & evaluator: KataGo~\citep{wu2019katago} \\
Chess         & board record (PGN), re-rendered              & $10^{8}$               & evaluator: Stockfish~\citep{stockfish} \\
Gomoku        & tournament record (PSQ), re-rendered         & $10^{4\,\dagger}$      & evaluator: Rapfi~\citep{rapfi} \\
Doudizhu      & self-play deal, rendered tabletop            & -                      & evaluator: PIMC~\citep{long2010pimc} \\
CounterStrike 2 & pro demo (\texttt{.dem}), radar render     & $10^{5\,\dagger}$      & parser: demo, 64\,Hz \\
DOTA~2        & Valve replay (\texttt{.dem}), minimap render & $10^{7}$               & parser: replay + API \\
StarCraft 2   & \texttt{.SC2Replay}, tiled top-down capture  & $10^{6\,\dagger}$      & parser: tracker events \\
GFootball     & self-play episode trace, rendered court      & -                      & parser: engine trace \\
Badminton     & stroke annotation, rendered court            & $10^{2}$               & parser: rollout \\
\midrule
\rowcolor{gray!15}
\multicolumn{4}{l}{\emph{Candidates: format and engine already open; only an adapter is missing.}}\\
Shogi         & kifu record (KIF/CSA), re-rendered           & $10^{6\,\dagger}$      & evaluator: YaneuraOu~\citep{yaneuraou} \\
Xiangqi       & game record, re-rendered                     & $10^{5\,\dagger}$      & evaluator: Pikafish~\citep{pikafish} \\
Othello       & WTHOR tournament archive, re-rendered        & $10^{5}$               & evaluator: Edax~\citep{edax} \\
LoL           & client replay (\texttt{.rofl})               & $10^{7\,\dagger}$      & parser: client API  \\
Hearthstone   & client power log, board render               & $10^{7\,\dagger}$      & parser: power log \\
AoE~II        & \texttt{.aoe2record}, tiled top-down capture & $10^{6\,\dagger}$      & parser: recorded game \\
Rocket League & community replay archive, broadcast render   & $10^{7\,\dagger}$      & parser: replay \\
$\vdots$      & \multicolumn{3}{l}{\emph{Hundreds more titles qualify: a replay format plus a data interface suffices.}}\\
\bottomrule
\end{tabular}%
}
\end{table}
\subsection{Task Definition and Notation}

Given a game environment and record, we can generate a video and recover the corresponding symbolic game states. Let $S_i$ denote the game states aligned with time step $i$. For each game, we define a game-related state variable $V_i = f(S_i)$ as a projection of $S_i$ that captures a key aspect of the evolving game world, such as win rate, economy balance, or match score. Depending on the game, $V_i$ is extracted directly from recorded state information or computed using a game-specific evaluator.
As the game evolves, we define a sequence of critical events $E = e_{1:N}$, such as moves in chess, or kills and tower destructions in MOBA games. Each event $e_j = (c_j, t_j, \mathcal{A}_j)$ is associated with an update to the game-related state variable  $V_{t_j}$, where $c_j$ denote the event category (what), $t_j$ the time step at which the event occurs (when), and $\mathcal{A}_j$ the set of involved entities (who).

We define three task categories to evaluate spatiotemporal world-state understanding and tracking. The model is expected to connect observations across time into a coherent understanding of how the game world changes. For each event $e_j$, we construct a reference tuple: $y_j=\bigl(e_j,\;u_j,\;s_j\bigr)$, whose three components correspond to three tasks. \textit{\textbf{Perceive (T1)}} requires identifying the event $e_j$ itself, i.e., what happened (the event category $c_j$), when (the timestamp $t_j$), and who was involved (the entity set $\mathcal{A}_j$). \textit{\textbf{Interpret (T2)}} requires predicting the state-update label $u_j$, which describes the change in $V$ associated with $e_j$ using a game-specific label space. \textit{\textbf{Track (T3)}} requires predicting the resulting state label $s_j$, which represents the state $V$ after the event $e_j$. For example, in chess, events are moves or captures; $V$ represents the win rate; $u_j$ indicates whether the move favors white, black, or neither, while $s_j$ indicates which side leads afterward or whether the position is balanced.


\subsection{Scalable Construction from Game Records}
Video-Arena introduces an automatic pipeline to convert game records into videos and aligned reference labels. As shown in Figure~\ref{fig:pipeline}, each game record is converted into three temporally aligned artifacts: an event stream, a state sequence, and a video with answer-bearing elements removed. The event $\{e_j\}_{j=1}^N$ and state variable $\{V_{t_j}\}_{j=1}^N$ sequence provide reference labels $\{y_j\}_{j=1}^{N}$ with the help of parsers and superhuman evaluators. They are the ground truth for the deterministic judgement of model predictions. A game-specific adapter handles event extraction, state reconstruction, and video generation. Once implemented, the adapter can process new tasks without human-written questions, manual annotations, or LLM judges. The vast and growing supply of game records support scalable construction of evaluation cases, while additional adapters extend coverage to new games with little additional efforts. 
Table~\ref{tab:flywheel} summarizes a list of candidate games and the available supply of their game records, which can usually be collected from game servers, sports and esports APIs, replay repositories, and demo archives, or generated through simulation environments. 

Game-specific adapters derive events and state labels through two mechanisms: (1) Use a parser to directly parse replay files or game engine traces to recover events and states, as in CS2, DOTA 2, SC2, GFootball and Badminton. (2) Use a parser to recover events from records and use an evaluator to derive the corresponding state labels, as in Go, Chess, Gomoku and Doudizhu. In both cases, the videos rendering is designed to preserve the visual evidence needed to interpret events and track state changes. We remove visual elements that directly reveal the target labels, such as scoreboard, messages, HUDs and subtitles. More details about data construction are in Appendix~\ref{app:envs}.

\begin{table}[t]
\centering
\caption{Video-Arena's gold set: \emph{Len}: video duration in minutes ($\text{mean}\pm\text{std}$). $|y|$: number of reference tuples per case ($\text{mean}\pm\text{std}$). Labels follow each environment's reference temporal granularity. Details on three-tier tasks and evaluator configurations are provided in Appendix~\ref{app:tiers}.}
\label{tab:games}
\adjustbox{max width=\linewidth}{%
\begin{tabular}{@{}l cc lll@{}}
\toprule
\textbf{Game} & \textbf{Len} & \textbf{$|y|$} & \textbf{T1: perceive event identity} & \textbf{T2: interpret state update} & \textbf{T3: track resulting state} \\
\midrule
\rowcolor{gray!15}
\multicolumn{6}{@{}l}{\emph{\textbf{Board}: Turn-based perfect-information grid and board games.}}\\
Gomoku    & $3.2\pm1.5$   & $40.4\pm8.8$  & move                                        & win-rate swing, Rapfi       & win-rate leader after \\
Go        & $30.3\pm17.0$ & $275.8\pm66.7$ & move, capture                               & win-rate swing, KataGo     & win-rate leader after \\
Chess     & $31.3\pm21.5$ & $88.1\pm23.1$ & move, capture                                & win-rate swing, Stockfish  & win-rate leader after \\
\midrule
\rowcolor{gray!15}
\multicolumn{6}{@{}l}{\emph{\textbf{Card}: Imperfect-information shedding card game with private hands.}}\\
Doudizhu  & $4.4\pm0.9$    & $21.5\pm4.3$  & move, type, card count & win-rate swing, PIMC & win-rate leader after \\
\midrule
\rowcolor{gray!15}
\multicolumn{6}{@{}l}{\emph{\textbf{FPS}: Tactical first-person shooter with round-based economy and objectives.}}\\
CS2       & $38.9\pm8.7$  & $186.0\pm34.3$ & kill, elimination, plant and etc. & round winner                & score pair after the round \\
\midrule
\rowcolor{gray!15}
\multicolumn{6}{@{}l}{\emph{\textbf{MOBA}: Multiplayer online battle arena with lane economy and teamfights.}}\\
DOTA~2    & $46.9\pm14.8$ & $29.9\pm11.0$  & kill, tower, barracks, Roshan and etc. & minute-level net-worth change & net-worth leader at interval end \\
\midrule
\rowcolor{gray!15}
\multicolumn{6}{@{}l}{\emph{\textbf{RTS}: Real-time strategy with macro-economy and base-building mechanics.}}\\
SC2       & $19.1\pm2.9$   & $67.9\pm22.3$  & battle, building destroyed                  & net resource-value swing    & resource-value leader after \\
\midrule
\rowcolor{gray!15}
\multicolumn{6}{@{}l}{\emph{\textbf{Sports}: Continuous real-time athletic simulations and ball-tracking.}}\\
GFootball & $25.0\pm6.1$    & $20.3\pm9.0$   & goal (scoring side, scorer)                 & goal-differential shift & score pair after the goal \\
Badminton & $19.9\pm2.5$   & $35.1\pm2.6$  & point (scoring side, landing zone)         & score-differential shift            & score pair after the point \\
\bottomrule
\end{tabular}%
}
\end{table}
\subsection{Golden Set Construction}
To facilitate affordable and reproducible evaluations, we curate a fixed \emph{golden set} of 180 tasks covering 9 representative games across 6 genres: board games \textit{(Go, Chess, Gomoku)}, card games \textit{(Doudizhu)}, FPS \textit{(CS2)}, MOBA \textit{(DOTA2)}, RTS \textit{(SC2)}, and Sports \textit{(GFootball, Badminton)}. Using the pipeline in Figure \ref{fig:pipeline}, we construct 20 tasks per game across three task categories: \textbf{\textit{T1}}--perceive event identity, \textbf{\textit{T2}}--interpret state update and \textbf{\textit{T3}}--track resulting state. Table~\ref{tab:games} shows the details of configurations in each game.
For example, in Doudizhu, models first perceive events, including moves, types and card counts, then estimate the win-rate swing associated with the event, and track the win-rate leader after the event. In DOTA2, models identify events such as first blood or teamfight, then interpret net-worth changes associated with the event in the minute-level and track which side leads in net worth at the interval end. Appendix~\ref{app:envs} provides the detailed task definitions for each game. Environments differ in required events and domain reasoning, tracking bottlenecks may arise from event recognition, state assessment, or temporal integration.


The golden set provides a fixed basis for comparison across systems, while the pipeline supports continued benchmark expansion. Additional records can be processed to generate private held-out cases or replace contaminated ones within several minutes and new game-specific adapters can extend coverage to further environments. Table~\ref{tab:flywheel} lists candidate game environments to demonstrate the extensibility, including Shogi, Xiangqi, Othello, Hearthstone. Moreover, we propose the golden set for affordable evaluation for the community: one Counter-Strike 2 task processed by GPT-5.6-Sol in Codex takes about 30 minutes, 74 tool calls and $\$$32, and evaluating nine models across 180 tasks under two protocols cost approximately $\$$18,000 in total at public list prices.


\subsection{Evaluation Protocols and Metrics}
\label{sec3:scoring}

\textbf{Evaluation Protocols.} We introduce two evaluation protocols for Video-Arena, namely \textbf{\textit{native visual}} and \textbf{\textit{agentic}} protocols. In visual protocol, the model receives the video and task instructions and directly outputs its predictions, without access to external media tools or video editing capabilities. This protocol measures \textbf{\textit{native visual spatiotemporal ability}}. In agentic evaluation, the model operate within an agent harness that supports frame extraction, code execution, and multi-turn analysis. This protocol measures \textbf{\textit{tool-assisted spatiotemporal ability}}, assessing whether tool use enables the model to better perceive events, interpret state updates, and track evolving states over time.

\textbf{Evaluation Metrics.} We use F1 scores in Video-Arena. Specifically, the evaluated model receives a video and task instructions, and finally predicts a sequence of tuples $\{\bar{y}_j\}_{j=1}^N$ containing \textbf{\textit{T1}}, \textbf{\textit{T2}}, and \textbf{\textit{T3}} labels. A predicted tuple $\bar{y}_j$ fully matches a reference tuple $y_j$ only when all non-temporal fields match exactly and the temporal field differs by at most $\pm3$ seconds. Predicted tuples are greedily matched one-to-one to the temporally nearest unused reference tuple with the same event identity. We report strict F1 score over fully matched tuples. This criterion requires models to correctly identify an event, interpret its associated state update, and track the resulting state to receive credit. 

\textbf{Diagnostic Metrics.} The F1 score normalized in $[0, 100]$ is the primary metric and we design tier-wise diagnostic scores to explain its failure modes. \textbf{\textit{T1}} localization can be decomposed by loosening the time tolerance, dropping entity identity, or evaluating event type separately, distinguishing \textbf{\textit{spatiotemporal precision}} from recognition. \textbf{\textit{T2}} and \textbf{\textit{T3}} can be evaluated after \textbf{\textit{T1}} time-matching while dropping the other fields, using macro accuracy over the state-update label $u_j$ and state-label $s_j$, distinguishing \textbf{\textit{spatiotemporal understanding}} from perception. By gradually relaxing timing tolerances and non-essential fields, we can pinpoint where performance drops. These diagnostic metrics complement the F1 metric, showing whether low scores stem from missed events, inaccurate timing, wrong entities, misjudged local impacts, or poor state tracking.

\begin{table*}[t]
    \centering
    \small
    \setlength{\tabcolsep}{6pt}
    \caption{F1 score in the visual protocol. Videos are sampled at 1 FPS by default, capped at 256 uniform frames for longer videos. \textbf{Bold} means the best performance; \underline{underline} means the second.}
    \label{tab:video_results}
    \adjustbox{max width=0.88\linewidth}{%
    \begin{tabular}{@{}l*{9}{c}r@{\hspace{0.2em}}l@{}}
    \toprule
    \multirow{2}{*}{\textbf{Model}} & \multicolumn{3}{|c|}{\textbf{Board}} & \multicolumn{1}{c|}{\textbf{Card}} & \multicolumn{1}{c|}{\textbf{FPS}} & \multicolumn{1}{c|}{\textbf{MOBA}} & \multicolumn{1}{c|}{\textbf{RTS}} & \multicolumn{2}{c|}{\textbf{Sports}} & \multicolumn{2}{c}{\multirow{2}{*}{\textbf{Overall}}} \\
    \cmidrule(lr){2-4}\cmidrule(lr){5-5}\cmidrule(lr){6-6}\cmidrule(lr){7-7}\cmidrule(lr){8-8}\cmidrule(lr){9-10}
     & \multicolumn{1}{|c}{Go} & Chess & \multicolumn{1}{c|}{Gomoku} & \multicolumn{1}{c|}{Doudizhu} & \multicolumn{1}{c|}{CS2} & \multicolumn{1}{c|}{DOTA~2} & \multicolumn{1}{c|}{SC2} & GFootball & \multicolumn{1}{c|}{Badminton} & \multicolumn{2}{c}{} \\
    \midrule
    \rowcolor{gray!15} \multicolumn{12}{c}{\textit{\textbf{Baselines}}} \\
    Random Tuple          & 0.0 & 0.1 & 0.0 & 0.0  & 0.0  & 0.0 & 0.2 & 0.0 & 0.0 & 0.0 & \\
    Set-time Random Tuple & 0.0 & 0.3 & 0.1 & 0.0  & 0.0  & 0.3 & 1.1 & 0.0 & 0.0 & 0.2 & \\
    Temporal Ceiling @256       & 87.8 & 83.5 & 100.0 & 100.0 & 79.5 & 69.5 & 99.8 & 93.6 & 98.2 & 90.2 & \\
    \rowcolor{gray!15} \multicolumn{12}{c}{\textit{\textbf{Proprietary Models}}} \\
    GPT-5.6-Sol           & \textbf{7.1}   & \textbf{47.3}   & \textbf{19.6}   & \underline{17.9} & \textbf{17.8}   & \textbf{2.2} & \textbf{3.4}   & \textbf{0.8}   & \textbf{2.9}   & \textbf{13.2} & \firstlogo \\
    Claude-Opus-5.0       & \underline{6.0} & \underline{19.7} & 10.6            & 16.2             & \underline{10.3} & 0.0          & 0.6            & \textbf{0.8}   & \underline{1.0} & \underline{7.2}  & \secondlogo \\
    GPT-5.6-Luna          & 2.8            & 13.0            & \underline{11.1} & 13.2             & 0.5             & 0.0          & \underline{1.4} & 0.0            & 0.3            & 4.7  & \\
    Gemini-3.7-Flash      & 1.7            & 11.8            & 6.1             & 16.5             & 5.0             & 0.2          & 0.6            & 0.0            & 0.3            & 4.7  & \\
    Gemini-3.1-Pro        & 2.2            & 11.5            & 8.7             & 14.2             & 2.1             & 0.1          & 0.9            & 0.2            & 0.3            & 4.5  & \\
    
    \rowcolor{gray!15} \multicolumn{12}{c}{\textit{\textbf{Open-source Models}}} \\
    Kimi K3               & 1.4            & 9.9             & 6.8             & \textbf{21.4}    & 1.8             & 0.4          & 0.0            & \textbf{0.8}   & 0.7            & 4.8  & \thirdlogo \\
    Qwen3.8-27B           & 1.1            & 4.0             & 3.2             & 12.3             & 0.6             & 0.3          & 0.0            & \underline{0.4} & 0.0            & 2.4  & \\
    MiniMax M3            & 0.9            & 4.3             & 2.1             & 10.2             & 1.1             & 0.2          & 0.1            & 0.2            & 0.1            & 2.1  & \\
    Qwen3.8-27B-no-think  & 0.3            & 1.3             & 1.0             & 5.5              & 0.0             & 0.1          & 0.1            & 0.0            & 0.2            & 0.9  & \\
    Qwen3.5-122B-A10B     & 0.1            & 0.6             & 0.8             & 5.3              & 0.1             & \underline{0.8} & 0.2         & 0.3            & 0.0            & 0.9  & \\
    \bottomrule
    \end{tabular}%
    }
\end{table*}

    \section{Experiments}
    \label{sec:expts}
    
    \subsection{Setup}
    
    \textbf{Models.}
    We evaluate nine models under both the visual and the agentic protocol, which gives ten evaluated configurations in each table: five proprietary models, i.e., GPT-5.6-Sol\ and GPT-5.6-Luna\citep{gpt56card}, Gemini~3.1~Pro~\citep{gemini31pro}, Gemini~3.7~Flash~\citep{gemini37flash}, and Claude-Opus-5.0~\citep{opus5}; and five open-source models, i.e., MiniMax-M3~\citep{minimaxm3}, Kimi~K3~\citep{kimik3report}, Qwen3.8-27B~\citep{qwen38} with and without its thinking mode, and Qwen3.5-122B-A10B~\citep{qwen35}; Qwen3.8-27B accounts for two of the five because both of its thinking modes are evaluated. We use different harnesses for different models: Codex~\citep{codex} with the GPT series, Antigravity~\citep{antigravity} with the Gemini series, and Claude Code~\citep{claudecode} with the Claude and open-source series. In the agentic protocol, model performance defaults to the model-harness pair.

    \textbf{Baselines.}
    We use 3 baselines. The first samples uniformly from the valid tuple vocabulary and video duration, matching the reference label count; the second uses the correct label count and  timestamps, while the remaining fields are uniformly sampled from the vocabulary; finally, temporal ceiling assumes oracle non-temporal predictions and considers an event recoverable only when a sampled frame falls within the $\pm3$s window. More detailed experiment setup is in Appendix~\ref{app:protocols}.


       \begin{table*}[t]
    \centering
    \small
    \setlength{\tabcolsep}{3.8pt}
    \caption{F1 score in the agentic protocol. Each case is restricted to 200 interaction turns and a 2-hour time limit. \textbf{Bold} means the best model-harness system performance; \underline{underline} means the second.}
    \label{tab:agentic_results}
    \adjustbox{max width=0.88\linewidth}{%
    \begin{tabular}{@{}lc*{9}{c}r@{\hspace{0.2em}}l@{}}
    \toprule
    \multirow{2}{*}{\textbf{Model}} & \multirow{2}{*}{\textbf{Harness}} & \multicolumn{3}{|c|}{\textbf{Board}} & \multicolumn{1}{c|}{\textbf{Card}} & \multicolumn{1}{c|}{\textbf{FPS}} & \multicolumn{1}{c|}{\textbf{MOBA}} & \multicolumn{1}{c|}{\textbf{RTS}} & \multicolumn{2}{c|}{\textbf{Sports}} & \multicolumn{2}{c}{\multirow{2}{*}{\textbf{Overall}}} \\
    \cmidrule(lr){3-5}\cmidrule(lr){6-6}\cmidrule(lr){7-7}\cmidrule(lr){8-8}\cmidrule(lr){9-9}\cmidrule(lr){10-11}
     & & \multicolumn{1}{|c}{Go} & Chess & \multicolumn{1}{c|}{Gomoku} & \multicolumn{1}{c|}{Doudizhu} & \multicolumn{1}{c|}{CS2} & \multicolumn{1}{c|}{DOTA~2} & \multicolumn{1}{c|}{SC2} & GFootball & \multicolumn{1}{c|}{Badminton} & \multicolumn{2}{c}{} \\
    \midrule
    \rowcolor{gray!15} \multicolumn{13}{c}{\textit{\textbf{Baselines}}} \\
    Random Tuple          & -- & 0.0 & 0.1 & 0.0 & 0.0 & 0.0 & 0.0 & 0.2 & 0.0 & 0.0 & 0.0 & \\
    Set-time Random Tuple & -- & 0.0 & 0.3 & 0.1 & 0.0 & 0.0 & 0.3 & 1.1 & 0.0 & 0.0 & 0.2 & \\
    \rowcolor{gray!15} \multicolumn{13}{c}{\textit{\textbf{Proprietary Models}}} \\
    Claude-Opus-5.0       & Claude Code  & \textbf{49.0}   & 63.2            & \underline{32.7} & \underline{29.1} & \textbf{84.5}   & \textbf{30.0}   & \textbf{14.6}   & \textbf{92.9}   & \textbf{63.2}   & \textbf{51.0} & \firstlogo \\
    GPT-5.6-Sol           & Codex        & 34.9            & \underline{71.5} & 29.1            & 21.1             & \underline{65.9} & \underline{15.9} & \underline{7.4}  & \underline{42.4} & \underline{24.5} & \underline{34.7} & \secondlogo \\
    Gemini-3.7-Flash      & Antigravity  & \underline{40.6} & \textbf{72.2}   & 30.5            & \textbf{31.5}    & 28.7            & 5.6             & 4.3             & 24.7            & 14.6            & 28.1 & \thirdlogo \\
    Gemini-3.1-Pro        & Antigravity  & 30.9            & 66.7            & \textbf{34.5}   & 19.1             & 2.4            & 5.9             & 0.3             & 1.3             & 0.4             & 17.9 & \\
    GPT-5.6-Luna          & Codex        & 14.2            & 57.4            & 15.1            & 6.8              & 4.0             & 1.8             & 3.1             & 0.4             & 1.2             & 11.6 & \\
    \rowcolor{gray!15} \multicolumn{13}{c}{\textit{\textbf{Open-source Models}}} \\
    Kimi K3               & Claude Code  & 27.7            & 54.6            & 18.2            & 19.6             & 27.3            & 9.6             & 4.5             & 7.9             & 3.7             & 19.2 & \\
    Qwen3.8-27B           & Claude Code  & 30.4            & 64.3            & 20.8            & 18.7             & 7.4             & 2.6             & 1.2             & 14.5            & 1.3             & 17.9 & \\
    MiniMax M3            & Claude Code  & 17.2            & 47.1            & 17.5            & 15.2             & 7.3             & 0.3             & 0.8             & 1.7             & 0.7             & 12.0 & \\
    Qwen3.8-27B-no-think  & Claude Code  & 18.5            & 26.3            & 15.2            & 16.3             & 0.1             & 4.9             & 0.5             & 0.7             & 0.2             & 9.2  & \\
    Qwen3.5-122B-A10B     & Claude Code  & 18.1            & 25.1            & 10.5            & 10.9             & 0.4             & 0.7             & 0.0             & 0.0             & 1.6             & 7.5  & \\
    \bottomrule
    \end{tabular}%
    }
    \end{table*}

    \begin{figure}[t]
    \centering
    \begin{minipage}[t]{0.43\linewidth}
        \vspace{0pt}
        \centering
        \includegraphics[width=\linewidth]{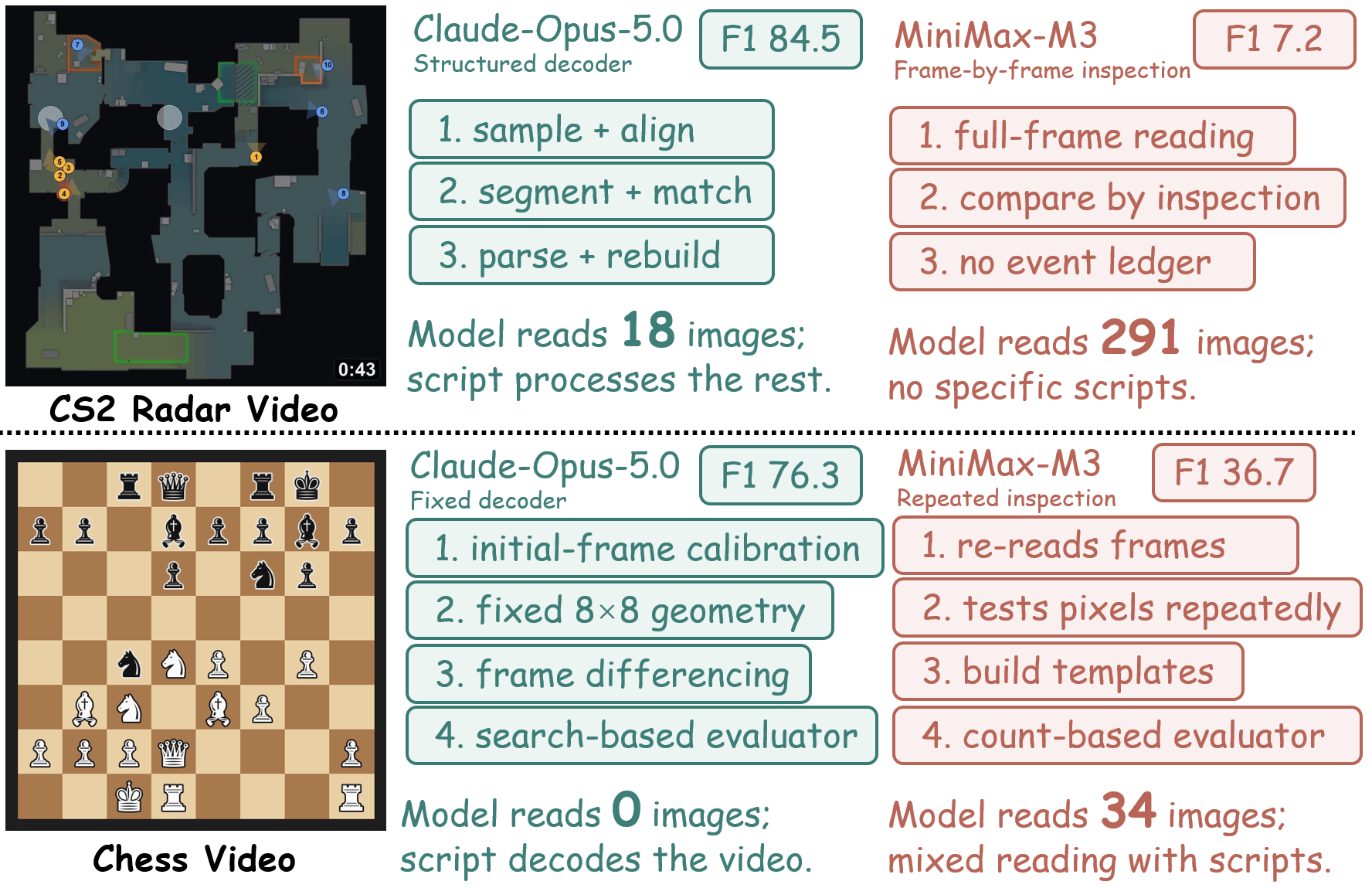}
    \end{minipage}\hfill
    \begin{minipage}[t]{0.52\linewidth}
        \vspace{0pt}
        \centering
        \setlength{\tabcolsep}{3pt}
        \adjustbox{max width=\linewidth}{%
        \begin{tabular}{@{}lccccccc@{}}
        \toprule
        \textbf{Setting} & \textbf{Board} & \textbf{Card} & \textbf{FPS} & \textbf{MOBA} & \textbf{RTS} & \textbf{Sports} & \textbf{Overall} \\
        \midrule
        \textit{Ceiling @256} & 90.4 & 100.0 & 79.5 & 69.5 & 99.8 & 95.9 & 90.2 \\
        \midrule
        \rowcolor{gray!15} \multicolumn{8}{c}{\textit{\textbf{GPT-5.6-Sol}}} \\
        A256     & 24.7 & 17.9 & 17.8 & 2.2 & 3.4 & 1.9 & 13.2 \\
        B256     & 37.8 & 34.0 & 27.1 & 9.6 & 3.6 & 2.1 & 21.3 \\
        BMP4     & 45.2 & 21.1 & \textbf{65.9} & \textbf{15.9} & \textbf{7.4} & 33.5 & 34.7 \\
        BMP4+256 & \textbf{48.2} & \textbf{36.4} & 46.6 & 6.3 & 6.5 & \textbf{48.6} & \textbf{37.5} \\
        \rowcolor{gray!15} \multicolumn{8}{c}{\textit{\textbf{GPT-5.6-Luna}}} \\
        A256     & 9.0 & 13.2 & 0.5 & 0.0 & 1.4 & 0.1 & 4.7 \\
        B256     & 23.5 & 13.6 & 5.0 & \textbf{6.4} & 0.0 & 1.0 & 10.9 \\
        BMP4     & 28.9 & 6.8 & 4.0 & 1.8 & \textbf{3.1} & 0.8 & 11.6 \\
        BMP4+256 & \textbf{31.9} & \textbf{14.9} & \textbf{20.9} & 3.0 & 0.6 & \textbf{12.7} & \textbf{17.8} \\
        \bottomrule
        \end{tabular}}
    \end{minipage}
    \caption{\textbf{Left}: Opus-5.0 and MiniMax-M3 solve two single cases using different methods in the agentic protocol. \textbf{Right}: The F1 score reports on four settings. A256 and BMP4 are visual and agentic protocol, while B256 replaces the video with 256 uniform frames in BMP4 and BMP4+256 provides additional 256 uniform frames on the basis of BMP4.}
    \label{fig:contrast}
    \end{figure}

    \subsection{Main findings: native vs. tool-assisted spatiotemporal ability}
    \label{sec:main_findings}

    \textbf{Native models struggle to perceive the evolving world.} With frames sampled in the video alone, GPT-5.6-Sol gets the best overall score $13.2$, while the mean across models is $4.6$ in Table~\ref{tab:video_results}. Models do better on board and card games, especially Chess, Gomoku and Doudizhu. Their performance drops sharply on continuous games: the best model reaches only $2.2$ on DOTA~2, $0.8$ on GFootball. Reliable event perception remains challenging even for games with relatively short videos or sparse event streams, such as Gomoku, GFootball, and Doudizhu. The difficulty is not simply watching more video, but interpreting and  tracking a world state that keeps changing.
    
    \textbf{Tools help event perception but remain insufficient for reliable state tracking.} The agentic evaluation protocol raises the mean overall score from $4.6$ to $20.9$, and Claude-Opus-5.0 reaches $51.0$ in Table~\ref{tab:agentic_results}. The gains are most pronounced when tools can turn visible cues into structured event records: Opus reaches $92.9$ on GFootball and $84.5$ on CS2. Yet this advantage becomes smaller when the task requires interpreting the accumulated situation: the same model reaches only $30.0$ on DOTA~2 and $14.6$ on SC2. Tools can help recover what happened; they do not by themselves tell the model what the changing world means.

    \textbf{Harness gains from structuring evidence, not simply adding images.} The best native multimodal model is not the best agentic model: GPT-5.6-Sol achieves the highest F1 score of $13.2$ under the native visual evaluation protocol. However, under agentic evaluation, Opus-5.0 improves from $7.2$ to $51.0$, while Gemini-3.7-Flash improves from $4.7$ to $28.1$. Figure~\ref{fig:contrast} (left) gives  two cases of CS2 and Chess. Opus-5.0 reads fewer images and builds more autonomous scripts than MiniMax-M3, gaining $12\times, 2\times$ on each case. An agent harness is most effective when it converts video into structured state representations that scripts can update and evaluate, rather than merely amplify the model's native spatiotemporal ability. Figure~\ref{fig:contrast} (right) compares four evaluation settings. Moving from A256 to B256 increases F1 by $8.1$ for GPT-5.6-Sol and $6.2$ for GPT-5.6-Luna, demonstrating the benefits of multi-turn tool use helps at fixed visual coverage.  Providing access to the full video instead of 256 sampled frames further raises GPT-5.6-Sol's F1 to $34.7$, while yielding only a small gain for Luna, indicating that the benefits of active evidence acquisition vary across models. Combining full-video access with sampled frames as visual context yields the highest overall scores, $37.5$ and $17.8$, though improvements are not consistent across all environment.

    \subsection{Why models fail: tier-wise diagnosis}
    \label{sec:failure_diagnosis}
    
\begin{figure}[t]
    \centering
    \includegraphics[width=0.92\linewidth]{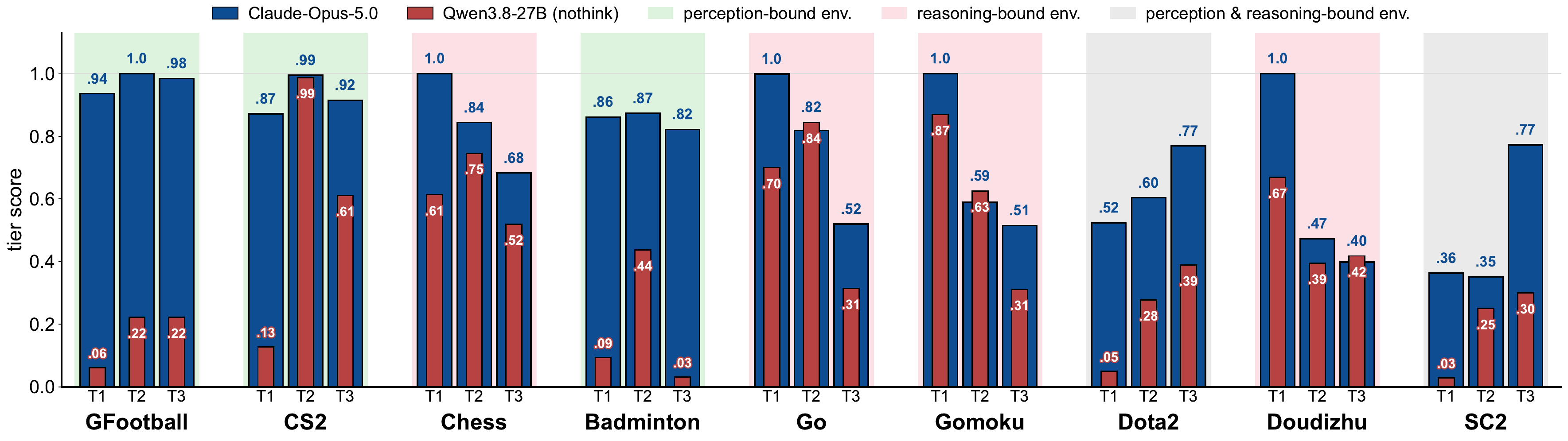}
    \caption{Tier scores for Claude-Opus-5.0 and Qwen3.8-27B-no-think on the shared valid cases of each environment. T1 is localization F1; T2 and T3 are accuracies over the events that T1 localized.}
    \label{fig:tiers_micro}
    \end{figure}

    We distinguish perceptual and reasoning bottlenecks by examining performance across the three task tiers (see Figure~\ref{fig:tiers_micro}). \textbf{\textit{T1}} measures event perception, while \textbf{\textit{T2}} and \textbf{\textit{T3}} evaluate state interpretation and tracking, respectively, on events correctly matched under \textbf{\textit{T1}}. We group environments according to whether the observed bottlenecks lie primarily in perception, reasoning, or both.
    
 \textbf{Perception-bound environments benefit most from reliable event extraction.} In GFootball, CS2 and Badminton, event recovery accounts for the gap between the two models. Board transitions and salient visual cues enable event extraction through scripts, but weaker models struggle to generate these scripts reliably. For score-based tasks, localized events largely determine the subsequent score update: Opus-5.0 reaches T3 accuracies of $0.98$ on GFootball and $0.92$ on CS2. These results indicate that recovering the event stream is often the main obstacle, rather than updating a counter.

\begin{figure}[t]
    \centering
    \includegraphics[width=0.92\linewidth]{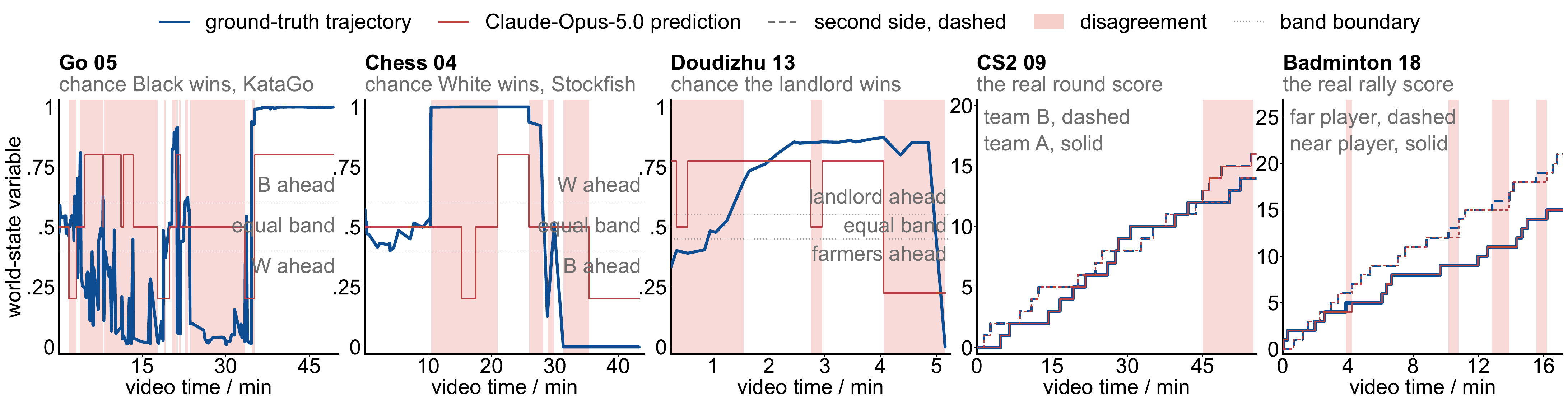}
    \caption{GT and predicted T3 trajectories for Claude-Opus-5.0 under the agentic protocol, with 5 separate cases. Score states track the reference closely, while win-rate states are difficult to track.}
    \label{fig:t3_trajectories}
    \end{figure}

  \textbf{Reasoning-bound environments remain difficult after events are recovered.} In Chess, Go, Gomoku and Doudizhu, Opus-5.0 achieves a T1 score of $1.00$, yet T3 accuracy reaches only $0.68$, $0.52$,
  $0.51$, and $0.40$, respectively. Recognizing every move does not ensure correct assessment of how it changes the position or who leads afterward. Majority-label predictions, especially ``equal'', further obscure errors on critical shifts. Figure~\ref{fig:t3_trajectories} illustrates the corresponding contrast: score trajectories track the reference closely, whereas win-rate states remain difficult to estimate.

  \textbf{Perception-and-reasoning-bound environments expose both bottlenecks across systems.} DOTA 2 and SC2 are challenging in both event perception and state inference, with the dominant bottleneck varying across model. Opus-5.0 reaches only $0.52$ and $0.36$ T1 on DOTA 2 and SC2. These groups summarize observed model bottlenecks rather than fixed intrinsic properties of the environments.

    \subsection{Frame-Budget Scaling and Agent-Turn Correlation}
    \label{sec:test_time_scaling}
\begin{figure*}[t]
    \centering
    \includegraphics[width=0.92\linewidth]{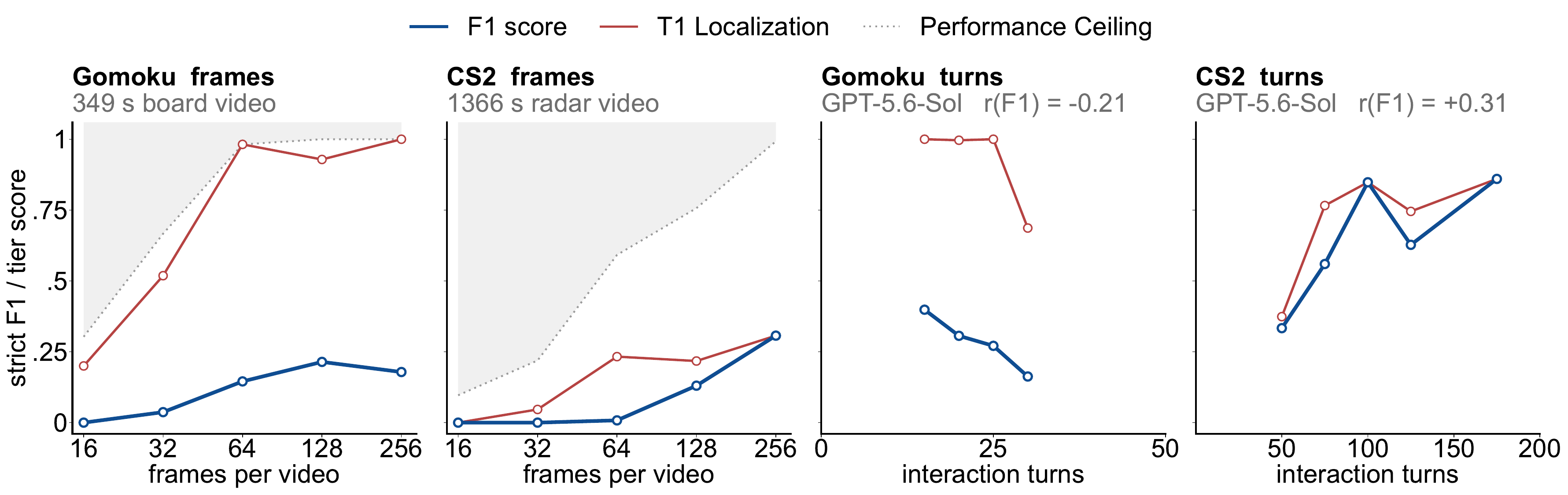}
    \caption{Test-time correlation analysis on GPT-5.6-Sol. The left two panels are sampled at $16$ to $256$ frames, with the performance ceiling imposed by frame spacing. The right two panels: complete-answer cases below the turn cap, pooled over and binned by turns. Gomoku spans $14$--$49$ turns in bins of $5$ turns, CS2 spans $62$--$200$ turns in bins of $25$. $r$: Pearson Correlation.}
    \label{fig:scaling}
    \end{figure*}

    We study two test-time budgets: sampled frames under the visual protocol, and interaction turns under the agentic protocol. Figure~\ref{fig:scaling} plots F1 and T1 location score against both.
    
\textbf{More frames improve perception rather than state understanding.} Because T1 requires a frame near each event, denser sampling can recover visual evidence that sparse sampling misses. On Gomoku, T1 rises from $20.0$ to $98.2$, while F1 score stops improving once localization is no longer the bottleneck. CS2 remains perception-limited at the same budget, so additional frames still help. Frame scaling improves event perception, but not the reasoning required to track the state.

\textbf{Turn use has environment-dependent performance implications, rather than acting as a universal accuracy dial.}  Turn count is an endogenous observation determined by the agent’s interaction trajectory, rather than a controlled budget. Among completed GPT-5.6-Sol runs, higher turn use is associated with scores rising from $33.3$ to $86.0$ on CS2, but falling from $39.8$ to $16.3$ on Gomoku. This contrast is consistent with their different task structures: CS2 rewards sustained sequential evidence gathering, whereas additional turns on Gomoku often indicate difficulty in reconstructing a single board state. Thus, the relationship between turn use and performance depends on the environment and should not be interpreted as a universal causal scaling law. This dependence is further illustrated by cross-system pooling, where capability differences reverse the Gomoku correlation.
    
    \subsection{Agentic behavior and protocol integrity}
    \label{sec:agentic_integrity}
    
    During early evaluation, we observed the following protocol-violating behaviors. We subsequently strengthened sandbox isolation and prohibited downloading external models, ensuring that the reported findings are reliable: \textbf{GT retrieval.} One GPT-5.6-Sol run accessed ground truth outside its sandbox and obtained a perfect score on an SC2 case, whereas the clean run reached only $0.058$ F1. \textbf{Evaluator substitution.} Two Qwen3.8-27B runs downloaded Stockfish to estimate chess values; these results show that external evaluators can distort the measurement without necessarily solving the underlying reconstruction problem. \textbf{Quit answering.} In about $5\%$ of visual-protocol cases, GPT-5.6-Luna declines to answer because it reasons that the protocol cannot yield a perfect score. This behavior does not indicate successful perception and is counted as a failure under the protocol.
    
\section{Conclusion and Limitation}
\label{sec:conclusion}

\textbf{Conclusion.} We introduced Video-Arena, turning spatiotemporal ability into a measurable ladder of competitive games, including event perception, state interpretation, and persistent tracking. Its vast data supply, deterministic judge, and the golden set provide scalable and comparable evaluations. Our experiments show that native models remain weak at integrating and keeping world state, while harness improves local evidence recovery without reliably interpreting value changes or persistently tracking the resulting state. Harness interaction alone does not remove this gap.

\textbf{Limitation.} We select 9 games with broad coverage, but they cannot exhaust the diversity of game worlds, and real-world states cannot always be converted into deterministic variables. Moreover, superhuman evaluators approximate but do not equal objective game-state variables; we mitigate this limitation by evaluating situation labels rather than exact win rates. Future work should extend the benchmark to broader environments, including real-world settings.

\subsection*{Ethics statement}

\textbf{Data sources.} All nine environments strictly comply with the agreements of their data sources. We reviewed the publicly available licences and terms of use of every upstream source, including the OGS, Lichess, and Gomocup game records, the DouZero, Google Research Football, and ShuttleSet releases, the public Counter-Strike~2 demos, the OpenDota match records, and Blizzard's StarCraft~II AI and machine learning licence. According to this review and the published terms, we use these data and all videos and annotations derived from them solely for scientific research, and this work involves no commercial activity. Game content remains the property of its publishers.

\clearpage
\bibliographystyle{unsrtnat}
\bibliography{main}

\begin{thebibliography}{59}
\providecommand{\natexlab}[1]{#1}
\providecommand{\url}[1]{\texttt{#1}}
\expandafter\ifx\csname urlstyle\endcsname\relax
  \providecommand{\doi}[1]{doi: #1}\else
  \providecommand{\doi}{doi: \begingroup \urlstyle{rm}\Url}\fi

\bibitem[Zwaan(1999)]{zwaan1999situation}
Rolf~A. Zwaan.
\newblock Situation models: The mental leap into imagined worlds.
\newblock \emph{Current Directions in Psychological Science}, 8\penalty0
  (1):\penalty0 15--18, 1999.

\bibitem[Zacks et~al.(2007)Zacks, Speer, Swallow, Braver, and
  Reynolds]{zacks2007event}
Jeffrey~M. Zacks, Nicole~K. Speer, Khena~M. Swallow, Todd~S. Braver, and
  Jeremy~R. Reynolds.
\newblock Event perception: A mind-brain perspective.
\newblock \emph{Psychological Bulletin}, 133\penalty0 (2):\penalty0 273--293,
  2007.
\newblock \doi{10.1037/0033-2909.133.2.273}.

\bibitem[Hasson et~al.(2008)Hasson, Yang, Vallines, Heeger, and
  Rubin]{hasson2008hierarchy}
Uri Hasson, Eunice Yang, Ignacio Vallines, David~J. Heeger, and Nava Rubin.
\newblock A hierarchy of temporal receptive windows in human cortex.
\newblock \emph{Journal of Neuroscience}, 28\penalty0 (10):\penalty0
  2539--2550, 2008.
\newblock \doi{10.1523/JNEUROSCI.5487-07.2008}.

\bibitem[Fu et~al.(2024)Fu, Dai, Luo, Li, Ren, Zhang, Wang, Zhou, Shen, Zhang,
  et~al.]{videomme}
Chaoyou Fu, Yuhan Dai, Yongdong Luo, Lei Li, Shuhuai Ren, Renrui Zhang, Zihan
  Wang, Chenyu Zhou, Yunhang Shen, Mengdan Zhang, et~al.
\newblock Video-mme: The first-ever comprehensive evaluation benchmark of
  multi-modal llms in video analysis.
\newblock \emph{arXiv preprint arXiv:2405.21075}, 2024.

\bibitem[Wu et~al.(2024)Wu, Li, Chen, and Li]{longvideobench}
Haoning Wu, Dongxu Li, Bei Chen, and Junnan Li.
\newblock Longvideobench: A benchmark for long-context interleaved
  video-language understanding, 2024.
\newblock URL \url{https://arxiv.org/abs/2407.15754}.

\bibitem[Chen et~al.(2024)Chen, Li, Dong, Zhang, Zang, Chen, Duan, Wang, Qiao,
  Lin, et~al.]{mmstar}
Lin Chen, Jinsong Li, Xiaoyi Dong, Pan Zhang, Yuhang Zang, Zehui Chen, Haodong
  Duan, Jiaqi Wang, Yu~Qiao, Dahua Lin, et~al.
\newblock Are we on the right way for evaluating large vision-language models?
\newblock \emph{arXiv preprint arXiv:2403.20330}, 2024.

\bibitem[Gu et~al.(2026)Gu, Xin, Zhang, Yang, Chua, Li, Zhang, Chen, Zhao, Xie,
  Liu, Lu, Han, Pavone, and Li]{vsisuperwild}
Tianjun Gu, Tianyu Xin, Kuan Zhang, Bowen Yang, Kok-Chung Chua, Peize Li,
  Xinran Zhang, Yupeng Chen, Qiyue Zhao, Qinlei Xie, Jianhang Liu, Yucheng Lu,
  Yinan Han, Marco Pavone, and Yiming Li.
\newblock Towards spatial supersensing in the wild.
\newblock In \emph{The Nineteenth European Conference on Computer Vision},
  2026.

\bibitem[Asadi et~al.(2026)Asadi, O'Sullivan, Cao, Nedaee, Rajabalifardi, Li,
  Adeli, and Ashley]{mirage}
Mohammad Asadi, Jack~W. O'Sullivan, Fang Cao, Tahoura Nedaee, Kamyar
  Rajabalifardi, Fei-Fei Li, Ehsan Adeli, and Euan Ashley.
\newblock Mirage: The illusion of visual understanding, 2026.
\newblock URL \url{https://arxiv.org/abs/2603.21687}.

\bibitem[Li et~al.(2024)Li, Guo, Guerin, and Lin]{dataconbination}
Yucheng Li, Yunhao Guo, Frank Guerin, and Chenghua Lin.
\newblock An open-source data contamination report for large language models.
\newblock In Yaser Al-Onaizan, Mohit Bansal, and Yun-Nung Chen, editors,
  \emph{Findings of the Association for Computational Linguistics: EMNLP 2024},
  pages 528--541, 2024.

\bibitem[Zhang et~al.(2025)Zhang, Chai, Xu, Zhang, Han, Yuan, Wang, and
  Cao]{noisylabel}
Kuan Zhang, Chengliang Chai, Jingzhe Xu, Chi Zhang, Han Han, Ye~Yuan, Guoren
  Wang, and Lei Cao.
\newblock Handling label noise via instance-level difficulty modeling and
  dynamic optimization.
\newblock \emph{Advances in Neural Information Processing Systems},
  38:\penalty0 52089–52118, 2025.

\bibitem[Chen et~al.(2026)Chen, Ye, Zhang, Du, Pu, Wang, Zuo, Duan, Ma, and
  Zhang]{chen2026snapbench}
Zirong Chen, Fuda Ye, Kuan Zhang, Enjun Du, Junfu Pu, Xinlei Wang, Xinyu Zuo,
  Lisheng Duan, Jin Ma, and Yongqi Zhang.
\newblock Snapbench: Benchmarking snap-and-ask multimodal retrieval for mobile
  interactions.
\newblock \emph{arXiv}, 2026.
\newblock URL \url{https://arxiv.org/abs/2608.29607}.

\bibitem[Yang et~al.(2026)Yang, Yang, Huang, Brown, Yang, Yu, Tong, Zheng, Xu,
  Wang, Lu, Fergus, LeCun, Fei-Fei, and Xie]{cambrians}
Shusheng Yang, Jihan Yang, Pinzhi Huang, Ellis Brown, Zihao Yang, Yue Yu,
  Shengbang Tong, Zihan Zheng, Yifan Xu, Muhan Wang, Daohan Lu, Rob Fergus,
  Yann LeCun, Li~Fei-Fei, and Saining Xie.
\newblock {Cambrian-S}: Towards spatial supersensing in video.
\newblock In \emph{International Conference on Learning Representations}, 2026.

\bibitem[Costarelli et~al.(2024)Costarelli, Allen, Hauksson, Sodunke,
  Hariharan, Cheng, Li, Clymer, and Yadav]{gamebench}
Anthony Costarelli, Mat Allen, Roman Hauksson, Grace Sodunke, Suhas Hariharan,
  Carlson Cheng, Wenjie Li, Joshua Clymer, and Arjun Yadav.
\newblock Gamebench: Evaluating strategic reasoning abilities of llm agents,
  2024.

\bibitem[Zhang et~al.(2026{\natexlab{a}})Zhang, Liu, Zhao, Hou, Zhang, Xie,
  Liu, and Li]{gameverse}
Kuan Zhang, Dongchen Liu, Qiyue Zhao, Jinkun Hou, Xinran Zhang, Qinlei Xie,
  Miao Liu, and Yiming Li.
\newblock Gameverse: Can vision-language models learn from video-based
  reflection?
\newblock \emph{arXiv preprint arXiv:2603.06656}, 2026{\natexlab{a}}.

\bibitem[Ouyang et~al.(2026)Ouyang, Hu, Lin, Ng, and Shou]{gameworld}
Mingyu Ouyang, Siyuan Hu, Kevin~Qinghong Lin, Hwee~Tou Ng, and Mike~Zheng Shou.
\newblock Gameworld: Towards standardized and verifiable evaluation of
  multimodal game agents, 2026.
\newblock URL \url{https://arxiv.org/abs/2604.07429}.

\bibitem[Tanaka and Simo-Serra(2021)]{lolv2t}
Tsunehiko Tanaka and Edgar Simo-Serra.
\newblock Lol-v2t: Large-scale esports video description dataset.
\newblock In \emph{Proceedings of the IEEE/CVF Conference on Computer Vision
  and Pattern Recognition (CVPR) Workshops}, pages 4552--4561, June 2021.

\bibitem[Zhang et~al.(2024)Zhang, Cao, Mo, Zhang, Poon, and Han]{gamemug}
Zhihao Zhang, Feiqi Cao, Yingbin Mo, Yiran Zhang, Josiah Poon, and Caren Han.
\newblock Game-mug: Multimodal oriented game situation understanding and
  commentary generation dataset, 2024.
\newblock URL \url{https://arxiv.org/abs/2404.19175}.

\bibitem[Ma et~al.(2026)Ma, Cao, Chen, Xu, Wang, and Jin]{egoesportsqa}
Jianzhe Ma, Zhonghao Cao, Shangkui Chen, Yichen Xu, Wenxuan Wang, and Qin Jin.
\newblock Egoesportsqa: An egocentric video benchmark for perception and
  reasoning in esports.
\newblock \emph{arXiv}, 2026.

\bibitem[Pan et~al.(2026)Pan, Yi, Ha, Islam, Zhang, Torresani, and
  Bertasius]{svibench}
Yulu Pan, Han Yi, Seongsu Ha, Md~Mohaiminul Islam, Benjamin Zhang, Lorenzo
  Torresani, and Gedas Bertasius.
\newblock Svi-bench: A dynamic microworld for strategic video intelligence.
\newblock \emph{arXiv}, 2026.

\bibitem[Jin et~al.(2026)Jin, He, Lo, Hu, and Rad]{seeingwithout}
Dingyang Jin, Jiawei He, Calvin Lo, Steven Hu, and Ryan Rad.
\newblock Seeing without understanding: Disentangling perception, reasoning,
  and simulation in {VLM} gameplay.
\newblock In \emph{Forty-third International Conference on Machine Learning},
  2026.

\bibitem[Zhang et~al.(2026{\natexlab{b}})Zhang, Liu, Zhao, Xin, Su, Wang, Yin,
  Ma, Li, Gu, et~al.]{gamesurvey}
Kuan Zhang, Dongchen Liu, Qiyue Zhao, Tianyu Xin, Yue Su, Haisheng Wang, Han
  Yin, Hongbo Ma, Peize Li, Tianjun Gu, et~al.
\newblock Towards generalist game players: An investigation of foundation
  models in the game multiverse.
\newblock \emph{arXiv preprint arXiv:2605.09965}, 2026{\natexlab{b}}.

\bibitem[OpenAI et~al.(2024)OpenAI, Hurst, Lerer, Goucher, and et~al.]{gpt4o}
OpenAI, Aaron Hurst, Adam Lerer, Adam~P. Goucher, and Adam~Perelman et~al.
\newblock Gpt-4o system card.
\newblock \emph{arXiv:2410.21276}, 2024.

\bibitem[Comanici et~al.(2025)Comanici, Bieber, Schaekermann, Pasupat, and
  et~al.]{gemini2.5}
Gheorghe Comanici, Eric Bieber, Mike Schaekermann, Ice Pasupat, and
  Noveen~Sachdeva et~al.
\newblock Gemini 2.5: Pushing the frontier with advanced reasoning,
  multimodality, long context, and next generation agentic capabilities.
\newblock \emph{arXiv:2507.06261}, 2025.

\bibitem[Bai et~al.(2025)Bai, Cai, Chen, and et~al.]{qwen3vl}
Shuai Bai, Yuxuan Cai, Ruizhe Chen, and Keqin~Chen et~al.
\newblock Qwen3-vl technical report.
\newblock \emph{arXiv:2511.21631}, 2025.

\bibitem[Cores et~al.(2024)Cores, Dorkenwald, Mucientes, Snoek, and
  Asano]{tvbench}
Daniel Cores, Michael Dorkenwald, Manuel Mucientes, Cees G.~M. Snoek, and
  Yuki~M. Asano.
\newblock Lost in time: A new temporal benchmark for videollms.
\newblock \emph{arXiv}, 2024.

\bibitem[Liu et~al.(2024)Liu, Li, Liu, Wang, Ren, Li, Chen, Sun, and
  Hou]{tempcompass}
Yuanxin Liu, Shicheng Li, Yi~Liu, Yuxiang Wang, Shuhuai Ren, Lei Li, Sishuo
  Chen, Xu~Sun, and Lu~Hou.
\newblock Tempcompass: Do video llms really understand videos?
\newblock \emph{arXiv}, 2024.

\bibitem[Yang et~al.(2025)Yang, Yang, Gupta, Han, Fei-Fei, and Xie]{model1}
Jihan Yang, Shusheng Yang, Anjali~W Gupta, Rilyn Han, Li~Fei-Fei, and Saining
  Xie.
\newblock Thinking in space: How multimodal large language models see,
  remember, and recall spaces.
\newblock In \emph{2025 IEEE/CVF Conference on Computer Vision and Pattern
  Recognition (CVPR)}, pages 10632--10643. IEEE, 2025.

\bibitem[Wang et~al.(2024)Wang, Zhang, Zohar, and Yeung-Levy]{model2}
Xiaohan Wang, Yuhui Zhang, Orr Zohar, and Serena Yeung-Levy.
\newblock Videoagent: Long-form video understanding with large language model
  as agent.
\newblock In \emph{European Conference on Computer Vision}, pages 58--76.
  Springer, 2024.

\bibitem[Feng et~al.(2025)Feng, Gong, Li, Guo, Wang, Peng, Wu, Zhang, Wang, and
  Yue]{model3}
Kaituo Feng, Kaixiong Gong, Bohao Li, Zonghao Guo, Yibing Wang, Tianshuo Peng,
  Junfei Wu, Xiaoying Zhang, Benyou Wang, and Xiangyu Yue.
\newblock Video-r1: Reinforcing video reasoning in mllms.
\newblock \emph{Advances in Neural Information Processing Systems},
  38:\penalty0 99114--99137, 2025.

\bibitem[Zhou et~al.(2024)Zhou, Shu, Zhao, Wu, Xiao, Yang, Xiong, Zhang, Huang,
  and Liu]{mlvu}
Junjie Zhou, Yan Shu, Bo~Zhao, Boya Wu, Shitao Xiao, Xi~Yang, Yongping Xiong,
  Bo~Zhang, Tiejun Huang, and Zheng Liu.
\newblock Mlvu: A comprehensive benchmark for multi-task long video
  understanding.
\newblock \emph{arXiv preprint arXiv:2406.04264}, 2024.

\bibitem[Yan et~al.(2026)Yan, Dai, Ran, Li, Lin, Jin, Xie, Liao, and
  Lian]{proact-vl}
Weicai Yan, Yuhong Dai, Qi~Ran, Haodong Li, Wang Lin, Tao Jin, Xing Xie, Hao
  Liao, and Jianxun Lian.
\newblock Proact-vl: A proactive videollm for real-time ai companions, 2026.
\newblock URL \url{https://arxiv.org/abs/2603.03447}.

\bibitem[Wu(2019)]{wu2019katago}
David~J. Wu.
\newblock Accelerating self-play learning in go.
\newblock \emph{arXiv preprint arXiv:1902.10565}, 2019.
\newblock URL \url{https://arxiv.org/abs/1902.10565}.

\bibitem[{Stockfish Developers}(2026)]{stockfish}
{Stockfish Developers}.
\newblock Stockfish, 2026.
\newblock URL \url{https://github.com/official-stockfish/Stockfish}.
\newblock Open-source chess engine; software citation, accessed 2026-09-04.

\bibitem[{Rapfi Developers}(2026)]{rapfi}
{Rapfi Developers}.
\newblock Rapfi, 2026.
\newblock URL \url{https://github.com/dhbloo/rapfi}.
\newblock Open-source Gomoku engine; software citation, accessed 2026-09-04.

\bibitem[Long et~al.(2010)Long, Sturtevant, Buro, and Furtak]{long2010pimc}
Jeffrey Long, Nathan~R. Sturtevant, Michael Buro, and Timothy Furtak.
\newblock Understanding the success of perfect information monte carlo sampling
  in game tree search.
\newblock In \emph{Proceedings of the AAAI Conference on Artificial
  Intelligence}, volume~24, pages 134--140, 2010.
\newblock \doi{10.1609/aaai.v24i1.7562}.
\newblock General determinization-based perfect-information Monte Carlo search
  paradigm.

\bibitem[{YaneuraOu Developers}(2026)]{yaneuraou}
{YaneuraOu Developers}.
\newblock Yaneuraou, 2026.
\newblock URL \url{https://github.com/yaneurao/YaneuraOu}.
\newblock Open-source shogi engine; software citation, accessed 2026-09-04.

\bibitem[{Pikafish Developers}(2026)]{pikafish}
{Pikafish Developers}.
\newblock Pikafish, 2026.
\newblock URL \url{https://github.com/official-pikafish/Pikafish}.
\newblock Open-source Xiangqi engine; software citation, accessed 2026-09-04.

\bibitem[{Edax Developers}(2026)]{edax}
{Edax Developers}.
\newblock Edax, 2026.
\newblock URL \url{https://github.com/abulmo/edax-reversi}.
\newblock Open-source Othello engine; software citation, accessed 2026-09-04.

\bibitem[{OpenAI}(2026{\natexlab{a}})]{gpt56card}
{OpenAI}.
\newblock {GPT-5.6} preview system card, June 2026{\natexlab{a}}.
\newblock URL \url{https://deploymentsafety.openai.com/gpt-5-6-preview}.
\newblock Official system card, published 26 June 2026; accessed 2026-07-31.

\bibitem[{Google DeepMind}(2026{\natexlab{a}})]{gemini31pro}
{Google DeepMind}.
\newblock Gemini 3.1 pro model card, February 2026{\natexlab{a}}.
\newblock URL \url{https://deepmind.google/models/model-cards/gemini-3-1-pro/}.
\newblock Official model card, published 19 February 2026; accessed 2026-07-31.

\bibitem[{Google DeepMind}(2026{\natexlab{b}})]{gemini37flash}
{Google DeepMind}.
\newblock Gemini 3.7 flash model card, August 2026{\natexlab{b}}.
\newblock URL
  \url{https://deepmind.google/models/model-cards/gemini-3-7-flash/}.
\newblock Official model card, published 13 August 2026; accessed 2026-08-31.

\bibitem[{Anthropic}(2026{\natexlab{a}})]{opus5}
{Anthropic}.
\newblock Introducing claude opus 5, July 2026{\natexlab{a}}.
\newblock URL \url{https://www.anthropic.com/news/claude-opus-5}.
\newblock Official release announcement, 24 July 2026; accessed 2026-07-31.

\bibitem[{MiniMax}(2026)]{minimaxm3}
{MiniMax}.
\newblock Minimax m3: Frontier coding, 1m context, native multimodality --- all
  in one model, June 2026.
\newblock URL \url{https://www.minimax.io/blog/minimax-m3}.
\newblock Official release blog, 1 June 2026; accessed 2026-07-31.

\bibitem[{Moonshot AI}(2026)]{kimik3report}
{Moonshot AI}.
\newblock Kimi k3 technical report, July 2026.
\newblock URL
  \url{https://github.com/MoonshotAI/Kimi-K3/blob/main/k3_tech_report.pdf}.
\newblock Released 27 July 2026 with the model weights; accessed 2026-07-31.

\bibitem[{Qwen Team}(2026{\natexlab{a}})]{qwen38}
{Qwen Team}.
\newblock {Qwen3.8}, 2026{\natexlab{a}}.
\newblock URL \url{https://qwen.ai/blog?id=qwen3.8}.
\newblock Official release announcement; accessed 2026-09-07.

\bibitem[{Qwen Team}(2026{\natexlab{b}})]{qwen35}
{Qwen Team}.
\newblock {Qwen3.5}: Towards native multimodal agents, February
  2026{\natexlab{b}}.
\newblock URL \url{https://qwen.ai/blog?id=qwen3.5}.

\bibitem[{OpenAI}(2026{\natexlab{b}})]{codex}
{OpenAI}.
\newblock Codex: official documentation, 2026{\natexlab{b}}.
\newblock URL \url{https://learn.chatgpt.com/docs}.
\newblock Accessed 2026-07-31.

\bibitem[{Google}(2026)]{antigravity}
{Google}.
\newblock Google antigravity: an agentic development platform, 2026.
\newblock URL \url{https://antigravity.google/}.
\newblock Accessed 2026-07-31.

\bibitem[{Anthropic}(2026{\natexlab{b}})]{claudecode}
{Anthropic}.
\newblock Claude code: official documentation, 2026{\natexlab{b}}.
\newblock URL \url{https://code.claude.com/docs/en/overview}.
\newblock Accessed 2026-07-31.

\bibitem[{Online Go Developers}(2026)]{onlinego}
{Online Go Developers}.
\newblock Online go, 2026.
\newblock URL \url{https://online-go.com}.
\newblock Online Go game-record ecosystem; source citation, accessed
  2026-09-04.

\bibitem[{Lichess Developers}(2026)]{lichess}
{Lichess Developers}.
\newblock Lichess, 2026.
\newblock URL \url{https://lichess.org}.
\newblock Open chess game-record service; source citation, accessed 2026-09-04.

\bibitem[Zha et~al.(2021)Zha, Xie, Ma, Zhang, Lian, Hu, and Liu]{douzero}
Daochen Zha, Jingru Xie, Wenye Ma, Sheng Zhang, Xiangru Lian, Xia Hu, and
  Ji~Liu.
\newblock Douzero: Mastering doudizhu with self-play deep reinforcement
  learning.
\newblock In \emph{Proceedings of the 38th International Conference on Machine
  Learning}, volume 139, pages 12333--12344. PMLR, 2021.
\newblock URL \url{https://arxiv.org/abs/2106.06135}.

\bibitem[{DouZero Baselines Contributors}(2026)]{douzero_weights}
{DouZero Baselines Contributors}.
\newblock Douzero baselines, 2026.
\newblock URL \url{https://huggingface.co/palemoky/douzero-baselines}.
\newblock Released DouZero-WP weights, accessed 2026-09-04.

\bibitem[{Valve Corporation}(2026)]{counterstrike2}
{Valve Corporation}.
\newblock Counter-strike 2, 2026.
\newblock URL \url{https://www.counter-strike.net}.
\newblock Game and demo ecosystem; source citation, accessed 2026-09-04.

\bibitem[{OpenDota Developers}(2026)]{opendota}
{OpenDota Developers}.
\newblock Opendota, 2026.
\newblock URL \url{https://www.opendota.com}.
\newblock Open DOTA 2 replay and match-data ecosystem; source citation,
  accessed 2026-09-04.

\bibitem[{Blizzard Entertainment}(2026)]{starcraft2}
{Blizzard Entertainment}.
\newblock s2client-proto, 2026.
\newblock URL \url{https://github.com/Blizzard/s2client-proto}.
\newblock StarCraft II replay and API protocol; source citation, accessed
  2026-09-04.

\bibitem[{Google Research}(2020)]{gfootball}
{Google Research}.
\newblock Google research football, 2020.
\newblock URL \url{https://github.com/google-research/football}.
\newblock Open-source football simulator; source citation, accessed 2026-09-04.

\bibitem[Wang et~al.(2023)Wang, Huang, Ik, and Peng]{shuttleset}
Wei-Yao Wang, Yung-Chang Huang, Tsi-Ui Ik, and Wen-Chih Peng.
\newblock Shuttleset: A human-annotated stroke-level singles dataset for
  badminton tactical analysis.
\newblock \emph{CoRR}, abs/2306.04948, 2023.
\newblock URL \url{https://arxiv.org/abs/2306.04948}.

\bibitem[{CoachAI Developers}(2026)]{coachai}
{CoachAI Developers}.
\newblock Coachai projects, 2026.
\newblock URL \url{https://github.com/wywyWang/CoachAI-Projects}.
\newblock Open-source badminton analysis and environment code; source citation,
  accessed 2026-09-04.

\end{thebibliography}

\clearpage
\beginappendix
\section{Benchmark and evaluation details}
\label{sec:appendix}

This appendix specifies the Video-Arena benchmark behind the main results. Appendix~\ref{app:envs} describes the nine environments, their source records, renderings, and ground-truth mechanisms. Appendix~\ref{app:tiers} gives the frozen three-tier targets and scoring conventions. Appendix~\ref{app:cost} summarizes selection, build cost, and evaluation aggregation. Appendix~\ref{app:protocols} gives the detailed evaluation protocols, model deployments, baselines, and metrics, including the human audit in Appendix~\ref{app:human}. Sandbox construction and protocol-integrity controls are described separately in Appendix~\ref{app:integrity}.

\subsection{Environments and data construction}
\label{app:envs}

The golden set contains 20 cases from each of nine environments: Go, Chess, Gomoku, Doudizhu, Counter-Strike~2, DOTA~2, StarCraft~II, GFootball, and Badminton. The cases cover board, card, FPS, MOBA, RTS, and sports regimes. Each case begins with a replay, record, engine trace, or simulator rollout. The corresponding adapter reconstructs the source state, emits objective events, computes the task-relevant state variable, and renders a synchronized video. The renderer removes answer-bearing HUD elements, labels, clocks, scoreboards, and other side channels while preserving the visual evidence needed to recover the event stream. All existing game data collected from the Internet strictly adheres to non-commercial principles, is used solely for academic research purposes, and is properly attributed to its original sources.

The golden set is a fixed measurement instrument rather than a claim that game data are scarce. The autonomous flywheel in Table~\ref{tab:flywheel} can generate additional held-out cases from the same adapters, refresh contaminated cases, and extend the benchmark to new environments. The cases are stratified by environment and trajectory shape before model evaluation; no model output is used to select or relabel them.

\textbf{Go.} Cases start from SGF records exported through the OGS public API for fixture game 27400000~\citep{onlinego} and are re-rendered as a plain $19\times19$ board. Source: \url{https://online-go.com} (accessed: 2026-09-04). Stones, move order, coordinates, names, clocks, capture counters, and evaluation bars are controlled by the rendering contract. The reference state is the Black win probability produced by KataGo~\citep{wu2019katago} with the pinned network and search configuration fixed with the benchmark. T1 identifies the move or capture, its time, colour, and board entities; T2 is the quantized change in win probability associated with that move; T3 is the win-rate leader after the move. Autonomous build time per sample is $\sim8$ minutes.

\textbf{Chess.} Cases start from the public Lichess export for fixture cZWtZEBE~\citep{lichess} and are re-rendered as a plain $8\times8$ board. Source: \url{https://lichess.org/game/export/cZWtZEBE} (accessed: 2026-09-04). The reference state is the White-oriented expected-score/WDL value produced by the pinned Stockfish software configuration~\citep{stockfish}. Draws map to the neutral state. T1 identifies the ply or capture, its time, moving colour, and destination; T2 quantizes the change in the White-oriented value associated with the ply; T3 reports the post-ply leader. Autonomous build time per sample is $\sim5$ minutes.

\textbf{Gomoku.} Cases start from competition PSQ records and are re-rendered as a plain $15\times15$ board. Source format documentation and the pinned evaluator are distributed with Rapfi~\citep{rapfi}; source: \url{https://github.com/dhbloo/rapfi} (accessed: 2026-09-04). Rapfi is used as the pinned evaluator; its raw side-to-move trajectory is converted to the tempo-neutral trajectory specified by the benchmark so that alternating move parity does not become a label shortcut. T1 identifies the placed stone and its coordinate; T2 is the quantized change in the tempo-neutral win probability associated with the move; T3 is the post-move leader. Autonomous build time per sample is $\sim2$ minutes.

\textbf{Doudizhu.} Cases are generated from DouZero-WP greedy self-play deals using the open DouZero environment and released weights~\citep{douzero,douzero_weights}; sources: \url{https://github.com/kwai/DouZero} and \url{https://huggingface.co/palemoky/douzero-baselines} (accessed: 2026-09-04). The evaluator uses hidden-hand determinization followed by adaptive 400--4000-world sampling and greedy rollouts with the released DouZero-WP policy, following the general sampling paradigm~\citep{long2010pimc}: farmer hands and ranks remain hidden from the evaluated prefix, while the landlord hand and bottom cards are face up. DouZero provides relevant DouDizhu self-play background, but the PIMC estimator itself is part of our benchmark implementation. This prevents the value target from collapsing to a visible card-count heuristic. T1 identifies the non-pass play, its player, combination type, and card count; T2 records the local win-rate change associated with the play; T3 records the landlord-versus-farmers leader after the play. Evaluator abstentions are represented as null tier labels and are excluded from the corresponding tier denominator and strict comparison, while the row remains a T1 event. Autonomous build time per sample is $\sim10$ minutes.

\textbf{Counter-Strike~2.} Cases begin with public professional-match demo (\texttt{.dem}) files published on HLTV, parsed at their native tick rate and re-rendered as top-down radar views. The demo format is the Valve demo ecosystem~\citep{counterstrike2}; the game source is \url{https://www.counter-strike.net} (accessed: 2026-09-04). Player positions, the bomb, map geometry, and event-relevant motion remain visible, while the HUD, scoreboard, round counter, kill feed, economy, names, and audio are removed; the retained bomb/countdown clock is the only textual timing aid. T1 identifies the type-specific event, time, and required side, player, or objective attributes. T2 is the winner of the round containing the event, i.e., the local round outcome associated with the round's sequence of events; T3 is the cumulative score after that round. Both are carried on the event record rather than exposed as separate per-event fields. Autonomous build time per sample is $\sim6$ minutes.

\textbf{DOTA~2.} Cases begin with Valve replays and OpenDota match records~\citep{opendota}, then are re-rendered on the real minimap terrain with ground-truth coordinates and game sprites. Sources: \url{https://www.opendota.com} and \url{https://github.com/odota} (accessed: 2026-09-04). The HUD, scoreboard, clock, net-worth graph, names, and objective counters are removed. The task-relevant state is the parsed net-worth advantage, not a learned win-probability model. T1 identifies one of six event types---first blood, teamfight, tower, barracks, Roshan, or ancient---with its type-specific attributes. T2 quantizes the change in net-worth advantage over the aligned minute-level interval associated with the event; T3 identifies the net-worth leader at that interval's end. Both tier labels are derived from the minute-level gold curve, use the fixed $\pm1000$ gold dead band, and are stored in the fields \texttt{minute\_gain} and \texttt{leader\_after}, respectively. Autonomous build time per sample is $\sim10$ minutes.

\textbf{StarCraft~II.} Cases begin with Blizzard replay files and are rendered as full-map isometric observer views rather than literal top-down views. The replay format and game source are documented by Blizzard's StarCraft~II API resources~\citep{starcraft2}; source: \url{https://github.com/Blizzard/s2client-proto} (accessed: 2026-09-04). Resource counters, production panels, minimaps, names, and other direct state readouts are removed. Events and resource-equivalent values are recovered from the replay tracker stream. T1 identifies the target battle or building-destruction event, its time, and coarse map region; T2 is the banded change in resource-equivalent value associated with the event; T3 is the banded resource-equivalent leader after the event. The value combines banked resources, army, economy, and technology. Autonomous build time per sample is $\sim20$ minutes.

\textbf{GFootball.} Cases are generated from randomized Google Research Football simulator episodes~\citep{gfootball} and rendered as synchronized full $1280\times720$, 24-FPS inclined 3D broadcast views in real time. Source: \url{https://github.com/google-research/football} (accessed: 2026-09-04). The reference state is the native score pair after each goal. T1 identifies the goal, its time, scoring side, scorer surname, and own-goal status; T2 is the four-class score-differential change associated with the goal, \texttt{taken}, \texttt{extended}, \texttt{levelled}, or \texttt{narrowed}; T3 is the exact score pair after the goal. The minimap and floating scorer-name label remain visible under the benchmark rendering contract. The corpus contains 20 cases; the same adapter can produce further held-out matches without human annotation. Autonomous build time per sample is $\sim3$ minutes.

\textbf{Badminton.} Cases use ShuttleSet stroke records~\citep{shuttleset} and a fixed-camera synthetic 3D renderer built from real-match trajectories. CoachAI-Projects is the upstream data and environment source~\citep{coachai}, not our renderer. Sources: \url{https://github.com/wywyWang/CoachAI-Projects} and \url{https://arxiv.org/abs/2306.04948} (accessed: 2026-09-04). The reference state is the native score pair after each point. T1 identifies the point, its time, scoring side, and terminal landing zone; T2 is the four-class score-differential change associated with the point; T3 is the exact score pair after the point. The fixed renderer is $1280\times720$ at 15 FPS and preserves player, racket, and shuttle motion while removing broadcast scoreboards, names, clocks, and annotations; it is a synthetic render, not a source replay. Autonomous build time per sample is $\sim15$ minutes.

\subsection{Frozen three-tier specification}
\label{app:tiers}

For every case, the reference is a sequence of tuples
\begin{equation}
 y_i=\bigl(e_i,u_i,s_i\bigr),
\end{equation}
where $e_i$ is T1, $u_i$ is T2, and $s_i$ is T3. T1 identifies the event, its timestamp, and the required entities. T2 describes the associated state update, and T3 reports the resulting task-relevant state. Labels follow each environment's reference temporal granularity: a move, play, or scoring transition; a round in CS2; an aligned minute-level interval in DOTA~2; or event resolution in SC2.

For ordinal value functions, the evaluator returns a scalar or probability trajectory. Fixed environment-specific dead bands map signed changes and resulting values to $\{\text{side A},\text{side B},\text{equal}\}$. For native score states, the target retains the score representation required by the environment. All thresholds, event vocabularies, aliases, and null-handling rules are versioned with the task specification and are applied identically to every system.

The tier instantiations are:
\begin{itemize}
    \item \textbf{Board games:} Go, Chess, and Gomoku use engine-derived win-probability trajectories. T1 identifies board transitions, T2 quantizes the local win-probability swing, and T3 reports the post-event leader.
    \item \textbf{Card game:} Doudizhu uses the hidden-hand PIMC trajectory. T1 identifies non-pass actions and their combination attributes, T2 reports the local impact, and T3 reports the post-event leader.
    \item \textbf{Observable competitive states:} CS2 uses round score, DOTA~2 uses parsed net-worth, and SC2 uses parsed resource or army value. Their T2 labels describe associated updates and their T3 labels describe the resulting state at the reference temporal granularity.
    \item \textbf{Sports:} GFootball and Badminton expose native score records. T1 identifies the scoring event and its visible entities, T2 describes the score transition, and T3 reports the exact score after the event.
\end{itemize}

\subsection{Scoring, selection, and cost}
\label{app:cost}

The headline metric is strict tuple-level F1. A predicted tuple matches a reference tuple only when all non-temporal fields agree and the timestamp falls within the fixed $\pm3$ second tolerance. Matching is greedy, one-to-one, and uses the nearest unused reference tuple with the same event identity. Tier-wise metrics are diagnostic projections of the same answers: T1 can be evaluated with controlled timing or entity relaxations, while T2 and T3 are evaluated after T1 time matching. These relaxed scores do not replace strict F1.

Per-case scores are averaged equally within each environment, and the overall score is the macro-average of the nine environment means. Missing tier labels do not enter that tier's denominator. Infrastructure failures are excluded from the reported model means and retained with their failure class; valid zeroes, format errors, budget-bound runs, and answers that never overwrite the seeded file remain valid outcomes. We report no interpolated values for missing cells.

The golden set has 180 cases in total. Its size is chosen to keep paired video and agentic comparisons auditable while preserving broad environment coverage. Build cost is dominated by evaluator prefix computation for Go, Chess, Gomoku, and Doudizhu, by replay parsing and rendering for CS2, DOTA~2, and SC2, and by simulator rendering for GFootball and Badminton. A representative expensive agentic case takes about 30 minutes and 74 tool calls at a cost of roughly \$32; full-panel costs therefore depend strongly on the selected systems and provider prices.

\subsection{Sample selection and corpus balance}
\label{app:selection}

\begin{table}[t]
\centering
\caption{Composition of the golden set. \emph{Len}: duration span of the case videos in minutes. \emph{Shape}: blowout / tense / reversal counts. \emph{Outcome}: which side ends the record ahead, over the 20 cases. StarCraft~II carries no shape label and is balanced on the leading side.}
\label{tab:selection}
\adjustbox{max width=\linewidth}{%
\begin{tabular}{@{}l ccc@{}}
\toprule
\textbf{Game} & \textbf{Len} & \textbf{Shape} & \textbf{Outcome} \\
\midrule
\rowcolor{gray!15}
\multicolumn{4}{@{}l}{\emph{\textbf{Board}: Turn-based perfect-information grid and board games.}}\\
Gomoku    & $1.3$--$7.0$   & $6/10/4$ & B $10$ / W $10$              \\
Go        & $12.0$--$75.2$ & $5/10/5$ & B $10$ / W $10$              \\
Chess     & $5.0$--$88.1$  & $3/11/6$ & W $8$ / B $8$ / draw $4$    \\
\midrule
\rowcolor{gray!15}
\multicolumn{4}{@{}l}{\emph{\textbf{Card}: Imperfect-information shedding card game with private hands.}}\\
Doudizhu  & $2.6$--$6.3$   & $7/7/6$  & farmers $14$ / landlord $6$ \\
\midrule
\rowcolor{gray!15}
\multicolumn{4}{@{}l}{\emph{\textbf{FPS}: Tactical first-person shooter with round-based economy and objectives.}}\\
CS2       & $22.8$--$54.6$ & $8/5/7$  & A $13$ / B $7$              \\
\midrule
\rowcolor{gray!15}
\multicolumn{4}{@{}l}{\emph{\textbf{MOBA}: Multiplayer online battle arena with lane economy and teamfights.}}\\
DOTA~2    & $20.3$--$67.5$ & $7/6/7$  & Dire $12$ / Radiant $8$     \\
\midrule
\rowcolor{gray!15}
\multicolumn{4}{@{}l}{\emph{\textbf{RTS}: Real-time strategy with macro-economy and base-building mechanics.}}\\
SC2       & $12.2$--$22.3$ & --       & A-led $10$ / B-led $10$     \\
\midrule
\rowcolor{gray!15}
\multicolumn{4}{@{}l}{\emph{\textbf{Sports}: Continuous real-time athletic simulations and ball-tracking.}}\\
GFootball & $15.0$--$35.0$ & $7/6/7$  & A $10$ / B $10$              \\
Badminton & $16.9$--$27.4$ & $6/7/7$  & A $9$ / B $11$              \\
\bottomrule
\end{tabular}%
}
\end{table}

Case selection is a curation step that runs on the source records before any model is evaluated, and it is balanced along three axes: the shape of the state trajectory, the distribution of the answers, and the duration of the case video. Per-environment quotas are fixed in advance, and the resulting composition is recorded with the benchmark so that the selection can be audited rather than taken on trust. Table~\ref{tab:selection} summarizes the curated corpus along these axes.

\textbf{Trajectory shape.} Every candidate is labelled from the winner-perspective trajectory of its state variable: a \emph{blowout} case is one in which the eventual winner holds the lead for most of the record, a \emph{reversal} case is one in which the lead changes hands and the eventual winner spends a substantial fraction of the record behind, and a \emph{tense} case is everything in between. The target is an even split, and the curated mix over the 160 labelled cases is 49 blowout, 62 tense, and 49 reversal. Doudizhu, DOTA~2, GFootball, and Badminton meet the quota exactly; the board environments skew toward tense cases, most strongly Chess at three, eleven, and six, since the shape of a recorded game can only be accepted or rejected, not set. StarCraft~II carries no shape label because its cases are engine-generated matches; it is balanced on the leading side instead, with ten A-led and ten B-led cases.

\textbf{Answer distribution.} Balance on the answers is enforced where selection can reach it, which is the case level: which side ends the record ahead. Go, Gomoku, GFootball, and StarCraft~II are exactly even at ten and ten, Chess is eight and eight with four draws, and Badminton is nine and eleven. Counter-Strike~2, DOTA~2, and Doudizhu keep the split of their source records at thirteen to seven, twelve to eight, and fourteen to six, because the winning side there is inherited from a replay that is selected as a whole.

\textbf{Duration coverage.} The case videos span $1.3$ to $88.1$ minutes, and every environment contributes a spread rather than a cluster. Where the record is generated rather than found, the coverage is set explicitly: the twenty GFootball cases are an arithmetic sweep from $15.0$ to $35.0$ minutes in steps of about one minute. Where the record is found, duration cannot be dialled, so cases are chosen to spread over the natural span of the environment instead of accumulating near its mode, which is why Go and Chess reach past an hour while Gomoku and Doudizhu stay under ten minutes. The spread is deliberate. Duration controls how much evidence a model has to hold at once, and a corpus concentrated at a single length would confound that factor with the environment.

\textbf{Provenance of the selection.} Each environment has a case manifest giving, for every case, its source identifier, trajectory shape, and duration, together with the thresholds used to assign that shape. The thresholds are environment-specific because the state variable is: Go, Chess, and Gomoku classify on the fraction of the record for which the eventual winner leads by more than a fixed win-probability margin, Counter-Strike~2 on a three-round deficit, DOTA~2 on an eight-thousand-gold deficit, and GFootball and Badminton are generated directly against the target quota. Candidate pools are larger than the curated sets, so selection is a choice rather than a relabelling of everything at hand: the twenty Counter-Strike~2 cases come from twenty-five rendered candidates and the twenty StarCraft~II cases from thirty engine matches. Within an environment the thresholds are fixed across cases and are never re-tuned for an individual case, as per-case tuning would make macro accuracy incomparable across the corpus. No model output enters any of these decisions.

\section{Evaluation protocols, models, baselines, and metrics}
\label{app:protocols}

This section gives the fixed conditions used for the two evaluation panels. Both protocols use the same 180 cases, task instructions, answer schema, temporal tolerance, deterministic judges, and aggregation rules. The only intended difference is whether the model receives a single sampled visual input or an agentic harness for gathering and organizing evidence.

\subsection{Evaluation protocols}

\textbf{Protocol A: native visual protocol.} Each case is delivered as one ordered sequence of frames with an explicit \texttt{video$\_$ts} array. The target sampling rate is 1 FPS. For videos longer than 256 seconds, the frames are uniformly distributed over the full video and capped at 256; shorter videos use the available 1-FPS sequence. The model receives no tools, code execution, external media inspection, replay, timing sidecar, ground truth, judge, second model, or additional interaction round. It returns one complete structured event list in a single request. The timestamp array supplied with the frames is authoritative.

\textbf{Protocol B: agentic protocol.} Each case is run in an opaque sandbox containing the full rendered \texttt{match.mp4}, \texttt{ffmpeg}, \texttt{ffprobe}, \texttt{instruction.md}, a pre-seeded \texttt{answer.json}, and the native tools supplied by the evaluated harness. The model may use frame extraction, video probing, scripts, image processing, OCR, and multi-turn analysis to inspect and organize the video. The deliverable is the final \texttt{answer.json}; prose in the final chat response is not scored. Each case has a maximum of 200 interaction turns and a hard wall-clock limit of 7200 seconds. The agent may not interact with or replay the game, access timing sidecars, value curves, ground truth, judges or self-checks, engine weights, external model downloads, or external channels that can retrieve the source match.

\textbf{The frozen prompts.} The instruction file carries the task itself: the event vocabulary, the coordinate or zone convention, the definition of the tracked variable and its dead bands, the row schema, and the $\pm3$ second matching rule. That part is byte-identical across the two protocols, so both protocols ask the same question. What differs is delivery, and only delivery. A sentence such as \emph{the answer file is already in the sandbox} is true under Protocol B and false under Protocol A, which is one request with no tools, no filesystem, and no second turn; sending it anyway is not a harmless mismatch, because a model told to incrementally write a file it cannot touch can answer with the empty skeleton the instruction prints. The delivery text is therefore marked inside the shared file rather than maintained as two files, each protocol renders only its own version, and a pre-flight check assembles the real prompt for all nine environments and rejects the panel before any tokens are spent if a Protocol A prompt still contains agentic delivery text, if the single-reply tail is missing, or if the tail does not name that environment's schema root key.

The two boxes below give both prompts as the model receives them. The Protocol~A box is one complete request. Its first part stands for the rendered instruction, which differs by environment in its event vocabulary, conventions, labels, and schema and runs to between about $4.5$k and $10.7$k characters; it is shown as a skeleton, and each line set between \texttt{<{}<} and \texttt{>{}>} summarises one part of that file rather than quoting it. From the heading \texttt{What you may use} to the end, the box is verbatim, and braces mark the only substitutions: \texttt{\{n\}} is the number of frames in the request, \texttt{\{dur\}} is the video duration to $0.1$~s, \texttt{\{dt\}} is the nominal frame spacing to $0.01$~s, and \texttt{video\_ts} lists the timestamps of all $n$ delivered frames to $0.01$~s, of which the box shows only the ends; the remaining braced fields take a fixed value per environment. The $n$ frames travel in the same request as separate images in temporal order, as described above. The Protocol~B box is the whole turn-one message under the 200-turn, 7200-second budget, reproduced from the archived runs; under that protocol the task reaches the agent only as the \texttt{instruction.md} it is told to read, rendered with its Protocol~B delivery text. For typesetting, both boxes print an em dash as \texttt{--} and a right arrow as \texttt{->}, print a run of blank lines as one, and mark with $\hookrightarrow$ the continuation of a line too long for the column; every other line break is the prompt's own.

The current version of the benchmark has since renamed all nine root keys to \texttt{events}, the key in the listings of Appendix~\ref{app:game_details}.

\begin{promptbox}{Protocol A --- one request as the model receives it. The \texttt{<{}<}\,\texttt{>{}>} lines summarise the environment's instruction; from \texttt{What you may use} on the text is verbatim, braces mark the substitutions, and only the \texttt{video\_ts} array is shortened.}
<< The title, e.g. "# Task: Reconstruct a Go game from a board-only video AND judge the
   strategy". >>

<< The input: what the rendered view shows (board, card table, radar, minimap, bird's-eye
   game world or match camera), what has been removed from it, and how an event's time is
   read off the video. >>

<< T1: the event types to report and the fields that identify each event, with the
   coordinate, seat, side or zone convention those fields use. >>

<< T2 and T3: the two labels every event carries, their definitions and fixed label sets,
   any dead band, and any cell that is left unscored. >>

<< The answer format: one JSON object whose single key `{root_key}` holds the event list,
   one row per event carrying `{time_field}` in video seconds and the fields above, with a
   worked example. >>

<< The scoring: each row is matched 1:1 to the nearest unused truth row with the same key
   and a time within the tolerance; strict F1 counts a row only if all of its fields are
   right; most files also define a T1-only F1 and T2/T3 macro accuracy, and several warn
   against constant-label shortcuts. >>

<< Notes: environment-specific cautions, e.g. that events are sparse and the whole video
   has to be scanned. >>

## What you may use

You are given the frames above and nothing else -- there is no {tool}, no code you can run, no file to write and no second turn. {label clause} by you. If you happen to recognise this {game|match}, do not answer from memory: an answer that does not follow from the frames is not what is being measured.

---
You are given the match as {n} frames, uniformly sampled from a {dur}-second video (one frame every {dt} video-seconds). Frame k (0-indexed) is at video_ts[k], in seconds:
video_ts = [0.00, {t_1}, {t_2}, ..., {t_(n-1)}]. Use these exact values for timestamps; do not recompute them.

This is one reply and there are no tools: no sandbox, no file to write, no second turn, and nothing runs after you stop. You may reason here first if it helps.

Your answer is the LAST ```json fenced block in this reply. It must hold the COMPLETE JSON
object with the single `{root_key}` key -- every row you want scored, in that one block.
Print no other JSON anywhere in the reply: no schema sketch, no partial dump, no repeat of
the answer. Nothing outside that last block is read.

<< plus the {n} frames, one image each, in temporal order >>
\end{promptbox}

\begin{promptbox}{Protocol B --- the whole turn-one message under the 200-turn, 7200-second budget. The task reaches the agent only as the \texttt{instruction.md} it is told to read.}
You are a HEADLESS autonomous agent -- there is no human to talk to.
Communicate ONLY by calling tools (running commands, reading and writing files); a text
reply is not an answer. Your one deliverable is the file `answer.json` in this folder.

Read `instruction.md` in this folder for the task and the exact `answer.json` schema. The only input is the video `match.mp4` in this folder. This is Windows + Git Bash: invoke the bundled tools as `./ffmpeg.exe` and `./ffprobe.exe` (a bare `ffmpeg` will not resolve). The video plays at 1:1 real speed, so a frame's video second IS its game second.

Solve it yourself, from the video. Do not use the network for anything: no downloading or installing packages, models or weights, no calls to any external service, API or model, and no looking this match up. `./ffmpeg.exe`, `./ffprobe.exe`, the shell, and the Python already installed are your only tools. Any network use voids the run, which is then scored zero whatever `answer.json` contains.

`./answer.json` is already in this folder -- it was created before your first turn, with the schema's root key and an empty list. Do not delete it, do not rename it, and do not change the root key. Overwrite it in place -- in full, valid JSON every time -- as soon as you have something to put in it, and keep overwriting it as you learn more. It must be valid JSON at every moment: rewrite it in full each time (read -> add -> write back), never leave it half-written. Each rewrite should contain everything the previous one had, plus the new rows; if you are deep into the video and the file is not getting longer, you are overwriting instead of appending.

Whatever is in `answer.json` when the run stops is what gets scored. If you are cut off,
every row already written still counts; rows kept only in your reasoning, to be written out at the end, score zero.

You have at most 200 tool-call turns and about 120 minutes of wall-clock time. When either limit is reached you are stopped immediately. Everything else -- including how you sample the video -- is up to you.
\end{promptbox}

\subsection{Reference evaluators and label stability}
\label{app:evaluators}

Four environments derive T2 and T3 from superhuman evaluators rather than directly recorded scores. Go, Chess, and Gomoku query a pinned engine at every move and express value from a fixed side's perspective; Gomoku additionally smooths adjacent evaluations to remove the side-to-move oscillation. Dou Dizhu uses a seeded PIMC estimator over hidden deals consistent with the landlord information set and abstains when a label remains too close to a band boundary. Table~\ref{tab:evaluator_params} records the default configurations and fixed quantization bands; the same metadata is stored with every case.

\begin{table}[!htbp]
\centering
\small
\caption{Default evaluator configurations. Bold values are the default settings; the other values define the sensitivity grid in Table~\ref{tab:evaluator_agreement}. Quantization bands remain fixed throughout.}
\label{tab:evaluator_params}
\begin{tabular}{@{}llll@{}}
\toprule
 & \textbf{Go} & \textbf{Chess} & \textbf{Gomoku} \\
\midrule
Engine & KataGo & Stockfish 18 & Rapfi \\
Network & \texttt{kata1-b18c384nbt} & NNUE (bundled) & \texttt{mix9svqfreestyle} \\
Interface & analysis engine & UCI & Gomocup protocol \\
Value $V$ & Black win rate & White expected score & Black win rate \\
 &  & $(w+d/2)/1000$ from WDL & $\sigma(\text{eval}/200)$ \\
Post-processing & --- & --- & trailing-2 move-pair mean \\
Search knob & \texttt{maxVisits} & \texttt{depth} & \texttt{turn\_ms} \\
Grid & 400 / \textbf{1600} / 6400 & 16 / \textbf{20} / 24 & 250 / \textbf{1000} / 4000 \\
Threads & 16 per position & 4 & 2 \\
Memory & --- & 512\,MB hash & 350\,MB max \\
Band & $0.10$ & $0.10$ & $0.10$ \\
\bottomrule
\end{tabular}

\vspace{0.6em}
\begin{tabular}{@{}ll@{}}
\toprule
\multicolumn{2}{@{}l}{\textbf{Dou Dizhu}} \\
\midrule
Estimator & PIMC over the landlord information set; greedy rollouts \\
Policy & DouZero-WP released weights, $\tau=0$ (greedy) \\
Value $V$ & landlord win probability = fraction of sampled deals won \\
Belief & uniform over count-consistent deals (no action-history inference) \\
Search knob & worlds per state, initial $\to$ adaptive cap \\
Grid & $100\to1000$ / $\mathbf{400\to4000}$ / $1600\to16000$ \\
Refinement & up to 6 rounds, pushing borderline rows past $2\sigma$ \\
Bands & $\varepsilon_{\text{T2}}=0.034$, $\varepsilon_{\text{T3}}=0.05$ \\
Seeding & one seed per case, recorded in the case file \\
Abstention & label within $2\sigma$ of its band edge $\to$ \texttt{null}, not scored \\
\bottomrule
\end{tabular}
\end{table}

\textbf{Sensitivity protocol.} We re-label all 20 cases per environment at the default setting (a repeat-run control), a weaker setting, and a stronger setting. Count-valued budgets use $\tfrac14\times$ and $4\times$ the default budget; Stockfish uses depths $16$, $20$, and $24$. Quantization bands are held fixed, so disagreement means that a scored T2 or T3 label changes. We report both raw agreement and Cohen's $\kappa$, since the T2 \texttt{equal} class is frequent in the board games.

\textbf{Results.} Repeat runs reproduce $98.6\%$ of T2 labels and $98.7\%$ of T3 labels overall ($7622/7726$ and $7659/7758$), with per-evaluator agreement of $96.4$--$100\%$ and $\kappa=0.886$--$1.000$. Under the weaker and stronger settings, the three board engines retain $94.9$--$99.0\%$ raw agreement. The small, non-monotone differences for Stockfish and Rapfi are comparable to their repeat-run variability, so we interpret these rows as a stability bound rather than evidence that a particular search budget is superior. The seeded PIMC control is exactly reproducible; its $4\times$ arm also agrees on every jointly labelled row. Its $\tfrac14\times$ arm has high conditional agreement but only $39.7\%$ T2 coverage because the noisier estimator abstains near band boundaries. Table~\ref{tab:evaluator_agreement} reports coverage explicitly to prevent this selective agreement from appearing stronger than it is. Overall, the discrete labels used for scoring are stable to re-execution and substantial changes in search effort.

\begin{table}[!htbp]
\centering
\small
\caption{Agreement with the default discrete labels under repeat, weaker, and stronger evaluator settings. Bands are fixed. \emph{Rows} counts jointly labelled events; \emph{Cov.} is the lower T2/T3 coverage relative to the default labels; \emph{agr.} is raw agreement. Dou Dizhu reports T3/T2 row counts because its estimator may abstain near band boundaries.}
\label{tab:evaluator_agreement}
\setlength{\tabcolsep}{4.5pt}
\begin{tabular}{@{}llrrrrrr@{}}
\toprule
\textbf{Evaluator} & \textbf{Setting} & \textbf{Rows} & \textbf{Cov.} & \textbf{T2 agr.} & \textbf{T2 $\kappa$} & \textbf{T3 agr.} & \textbf{T3 $\kappa$} \\
\midrule
KataGo (Go) & 1600 visits, rerun (control) & 5138 & 100.0 & 99.2 & 0.964 & 99.3 & 0.988 \\
 & 400 visits ($\tfrac14\times$) & 5138 & 100.0 & 98.0 & 0.916 & 98.6 & 0.975 \\
 & 6400 visits ($4\times$) & 5138 & 100.0 & 98.7 & 0.941 & 99.0 & 0.982 \\
\midrule
Stockfish (Chess) & depth 20, rerun (control) & 1430 & 100.0 & 97.8 & 0.886 & 97.6 & 0.962 \\
 & depth 16 ($-4$) & 1430 & 100.0 & 96.8 & 0.839 & 96.6 & 0.946 \\
 & depth 24 ($+4$) & 1430 & 100.0 & 97.6 & 0.876 & 96.1 & 0.938 \\
\midrule
Rapfi (Gomoku) & 1000\,ms/move, rerun (control) & 808 & 100.0 & 96.4 & 0.905 & 96.4 & 0.942 \\
 & 250\,ms/move ($\tfrac14\times$) & 808 & 100.0 & 94.9 & 0.863 & 96.0 & 0.936 \\
 & 4000\,ms/move ($4\times$) & 808 & 100.0 & 96.2 & 0.898 & 97.0 & 0.952 \\
\midrule
PIMC (Dou Dizhu) & $400\!\to\!4000$, rerun (control) & 382/350 & 100.0 & 100.0 & 1.000 & 100.0 & 1.000 \\
 & $100\!\to\!1000$ ($\tfrac14\times$) & 338/139 & 39.7 & 99.3 & 0.988 & 100.0 & 1.000 \\
 & $1600\!\to\!16000$ ($4\times$) & 380/349 & 99.5 & 100.0 & 1.000 & 100.0 & 1.000 \\
\bottomrule
\end{tabular}
\end{table}

\subsection{Models, deployments, and concurrency}

We evaluate ten system rows comprising seven models accessed through their official APIs and three Qwen systems deployed on our own infrastructure. The official-API group contains GPT-5.6-Sol, GPT-5.6-Luna, Gemini-3.1-Pro, Gemini-3.7-Flash, Claude-Opus-5.0, Kimi K3, and MiniMax M3. The Qwen group contains Qwen3.8-27B, its no-thinking configuration, and Qwen3.5-122B-A10B. The Qwen systems run on vLLM with eight H200 GPUs. We use concurrency five per system for Protocol A and concurrency three per system for Protocol B. These are fixed scheduling conditions for comparable measurement, not additional model-time or quality budgets.

\textbf{Reasoning effort.} Throughout the evaluation, we hold the reasoning effort fixed for each model across all nine environments, ensuring comparability across environments within each row. Specifically, we set the reasoning effort to \texttt{high} for GPT-5.6-Sol, GPT-5.6-Luna, Claude-Opus-5.0, Gemini-3.1-Pro, and Gemini-3.7-Flash. For MiniMax-M3 and Kimi-K3, we retain the provider-default reasoning effort. For the two Qwen models, reasoning is governed via the \texttt{enable$\_$thinking} parameter.

\subsection{Random baselines}

Both random controls are generated and scored offline without a model, video access, or inclusion in the model aggregate. For each case, the predicted number of rows equals the number of reference rows. \textit{Random Tuple} samples every answer field, including time, from the task's declared legal vocabulary or numeric range and samples time within the video duration. It does not estimate a distribution from reference values. \textit{Set-time Random Tuple} reuses the reference timestamp for the corresponding row but samples every non-temporal field independently from the same task-level vocabulary or range. It therefore measures the chance performance of knowing when an event occurred while not knowing its content or consequences.

\subsection{Temporal sampling ceiling}
\label{app:ceiling}

Protocol A delivers a fixed set of uniformly spaced still frames rather than a continuously decodable video, so part of the temporal information the judge asks about is destroyed before the model sees anything, and the third baseline row of Table~\ref{tab:video_results} measures exactly how much. A case of duration $D$ is delivered as $n=\min(256,\lfloor D\rfloor)$ uniform frames, so the frame spacing is $\Delta=D/n$, which equals one second whenever the 1-FPS sequence fits inside the cap and grows linearly with duration once it does not. The delivered timestamps are the grid $T=\{0,\Delta,2\Delta,\dots,(n-1)\Delta\}$; the prompt lists $T$ verbatim and instructs the model to use those values rather than recompute them, so $T$, and not the underlying video, is the set of times an answer may name. A model with flawless perception still learns only which $\Delta$-wide bin an event fell into: every event interior to a bin produces the same frame pair, so no amount of visual ability distinguishes them. Its best strategy is to answer the bin it saw, and that answer is time-matched only when the reported timestamp lies within the tolerance $\tau=3$~s of the true event.

\textbf{Uniform temporal ceiling.} If event times are assumed uniformly distributed with respect to the grid, some grid point falls within $\tau$ of a given event for a fraction
\begin{equation}
c \;=\; \min\!\left(1,\; \frac{2\tau}{\Delta}\right)
\label{eq:ceiling}
\end{equation}
of events. The bound is vacuous while $\Delta\le 2\tau=6$~s --- that is, for every video shorter than $1536$~s --- and binds only above it. Equation~\ref{eq:ceiling} is a \emph{recall} quantity: it counts the reference events the grid can time at all. An oracle whose non-temporal fields are all correct then has two strategies. Answering every reference event gives $P=R=c$ and hence $F1=c$. Abstaining on the events it cannot time gives $P=1$, $R=c$, and
\begin{equation}
F1_{\text{unif}} \;=\; \frac{2c}{1+c},
\label{eq:ceiling_f1_unif}
\end{equation}
which dominates the first strategy for every $c<1$. Since the headline metric is strict tuple $F1$, Equation~\ref{eq:ceiling_f1_unif} rather than Equation~\ref{eq:ceiling} is the comparable bound, and quoting $c$ in an $F1$ table understates the available headroom.

\textbf{Actual temporal ceiling.} Equation~\ref{eq:ceiling} still replaces the real event times with a uniformity assumption. The actual ceiling drops that assumption and reads the reference file of each case. For a case with reference set $G$, $N=|G|$, and delivered grid $T$, let
\begin{equation}
M \;=\; \bigl|\{\, g\in G \;:\; \exists\, t\in T,\; |t-t_g|\le\tau \,\}\bigr|
\label{eq:ceiling_m}
\end{equation}
be the number of reference events the grid can time, and $R_{\text{act}}=M/N$. The same abstaining oracle attains $TP=M$, $FP=0$, $FN=N-M$, so
\begin{equation}
F1_{\text{act}} \;=\; \frac{2M}{N+M} \;=\; \frac{2R_{\text{act}}}{1+R_{\text{act}}}.
\label{eq:ceiling_f1_act}
\end{equation}
We compute Equation~\ref{eq:ceiling_f1_act} per case from the case's own grid and reference times, average over the 20 cases of an environment, and macro-average over the nine environments. This is the quantity reported as \textit{Temporal Ceiling @256} in Table~\ref{tab:video_results} and as \textit{Ceiling @256} in Figure~\ref{fig:contrast}; the oracle may name any delivered timestamp, which is exactly the freedom the prompt grants.

\textbf{The two ceilings differ for two reasons.} The first is phase. Real events are not uniformly placed relative to the grid, and the deviation runs in both directions: on the worst case of the corpus, GFootball \texttt{gf-20}, Equation~\ref{eq:ceiling_f1_unif} reads $84.5$ while only two of six reference events are reachable and $F1_{\text{act}}=50.0$; on DOTA~2 \texttt{dota-15} the real placement is luckier than uniform, $59.8$ against $68.5$. The second is the tail. The grid spans $[0,(n-1)\Delta]=[0,D-\Delta]$ and therefore stops one full interval before the video ends, so an event in the final $\Delta$ seconds is unreachable even where $\Delta\le2\tau$ predicts that everything is reachable. Of the $3{,}279$ unreachable reference events in the corpus, $210$ lie past the last delivered frame, and in $34$ of the $180$ cases the uniform bound reads exactly $100$ while the actual ceiling does not --- in every one of those $34$ cases the entire loss is tail loss. Badminton is the clearest instance: a uniform ceiling of $99.8$, with $19$ of its $20$ cases at exactly $100$, against an actual ceiling of $98.2$ with all $20$ cases bitten by the tail. Aggregated, the two effects nearly cancel ($91.5$ against $90.2$ overall), but they do not cancel per case, and only the actual ceiling is a bound on any particular cell.

Table~\ref{tab:ceiling_compare} reports both quantities for all nine environments alongside the recall bound $c$ that they are derived from. The measured spacings make the row read as a property of video length rather than of game genre. Gomoku and Doudizhu render short enough for a full 1-FPS sequence (median $\Delta$ of $1.00$~s and $1.03$~s) and lose nothing; StarCraft~II, at a median $4.58$~s, stays under the analytic bound and loses only $6$ events, all of them in the tail. Badminton ($4.47$~s) and GFootball ($5.86$~s) reach $98.2$ and $93.6$; Go ($6.21$~s) and Chess ($7.16$~s) fall to $87.8$ and $83.5$; the two longest environments cost the most, Counter-Strike~2 at a median $9.13$~s giving $79.5$ and DOTA~2 at $12.16$~s giving $69.5$. Averaged over the nine environments the actual ceiling is $90.2$.

\begin{table}[!htbp]
\centering
\small
\caption{Uniform against actual temporal ceiling at the 256-frame grid, $\tau=3$~s, over all $180$ cases and $15{,}278$ reference events. $c$ is the uniform recall bound of Equation~\ref{eq:ceiling}; $F1_{\text{unif}}$ is its $F1$ form (Equation~\ref{eq:ceiling_f1_unif}); $F1_{\text{act}}$ is the actual ceiling computed from the real event times (Equation~\ref{eq:ceiling_f1_act}) and is the value reported in the main tables. \emph{Unreached} counts reference events no delivered timestamp can match, and \emph{tail} is the subset of those lying past the last delivered frame. All ceilings are per-case means on the $[0,100]$ scale.}
\label{tab:ceiling_compare}
\adjustbox{max width=\linewidth}{%
\begin{tabular}{@{}lrrrrrrr@{}}
\toprule
\textbf{Environment} & \textbf{Median $\Delta$ (s)} & \textbf{Events $N$} & \textbf{$c$} & \textbf{$F1_{\text{unif}}$} & \textbf{$F1_{\text{act}}$} & \textbf{Unreached} & \textbf{Tail} \\
\midrule
Go               & 6.21  & 5{,}516 & 82.5  & 88.4  & 87.8  & 1{,}128 & 44 \\
Chess            & 7.16  & 1{,}761 & 78.5  & 85.8  & 83.5  & 494     & 49 \\
Gomoku           & 1.00  & 808     & 100.0 & 100.0 & 100.0 & 0       & 0  \\
Doudizhu         & 1.03  & 429     & 100.0 & 100.0 & 100.0 & 0       & 0  \\
Counter-Strike~2 & 9.13  & 3{,}703 & 68.8  & 80.6  & 79.5  & 1{,}288 & 69 \\
DOTA~2           & 12.16 & 597     & 59.7  & 72.6  & 69.5  & 299     & 20 \\
StarCraft~II     & 4.58  & 1{,}358 & 100.0 & 100.0 & 99.8  & 6       & 6  \\
GFootball        & 5.86  & 405     & 92.7  & 96.0  & 93.6  & 39      & 2  \\
Badminton        & 4.47  & 701     & 99.7  & 99.8  & 98.2  & 25      & 20 \\
\midrule
\textbf{Overall} &       & 15{,}278 & \textbf{86.9} & \textbf{91.5} & \textbf{90.2} & 3{,}279 & 210 \\
\bottomrule
\end{tabular}%
}
\end{table}

The ceiling is an upper bound on what our own sampling permits and not a prediction of any model's score: it assumes every non-temporal field is already correct, so field errors, missed events, and hallucinated rows all subtract from it, and a model may legitimately land far below it. It is also protocol-specific. Under Protocol B the agent chooses its own frame spacing with \texttt{ffmpeg}, so $\Delta$ is a decision the agent makes rather than one we impose and the ceiling is $1$ by construction; Table~\ref{tab:agentic_results} therefore carries no such row. Two variants bracket the reading of Equation~\ref{eq:ceiling_m}. Restricting the oracle to name only a frame at or after the event, as a strictly causal reader would, halves the admissible window and gives an overall ceiling of $72.2$; allowing it to interpolate the midpoint of the bracketing pair, which the prompt does not permit, gives $91.2$. We report the unrestricted variant because the prompt fixes the answerable timestamps to the delivered grid. Every cell records its own $n$, effective frame rate, and ceiling in the run record next to its score, so the quantity is auditable per case rather than reconstructed from the table; a cell whose strict $F1$ approaches its own ceiling with the residual misses concentrated in correctly typed but mistimed rows is reported as a sampling-limited outcome, and the ceiling is quoted with it.

\subsection{Metrics and aggregation}
\label{app:metrics}
Let $G$ be the reference tuple set and $P$ the predicted tuple set. Under greedy one-to-one matching, a pair contributes to $TP$ only when every non-temporal field is correct and the time error is at most 3 seconds. Unmatched predictions are $FP$ and unmatched references are $FN$. We compute
\begin{equation}
 P=\frac{TP}{TP+FP},\qquad
 R=\frac{TP}{TP+FN},\qquad
 F1=\frac{2PR}{P+R},
\end{equation}
with the zero-denominator cases defined as zero. The strict tuple-level $F1$ is the headline score and is reported as a percentage.

T1 is the event-localization and grounding score. Its matched units require the event identity, timestamp, and task-required entities under the specified temporal tolerance; controlled relaxed variants can remove one of these requirements for diagnosis. T2 and T3 are conditional scores on events correctly localized by T1, not scores on an easier subset selected by the predicted tier labels. Let $M_{T1,k}$ be the one-to-one T1-matched events with a non-null reference label for tier $k$, and let $\hat y_{i,k}$ and $y_{i,k}$ be the predicted and reference labels. For $k\in\{2,3\}$, we compute
\begin{equation}
 \operatorname{Acc}_{k\mid T1}=\frac{1}{|M_{T1,k}|}
 \sum_{i\in M_{T1,k}}\mathbf{1}\!\left[\hat y_{i,k}=y_{i,k}\right].
\end{equation}
For categorical labels, the headline tier diagnostic is macro accuracy over the valid label set $\mathcal{C}_k$:
\begin{equation}
 \operatorname{MacroAcc}_{k\mid T1}=\frac{1}{|\mathcal{C}_k|}
 \sum_{c\in\mathcal{C}_k}\frac{\sum_{i\in M_{T1,k}}\mathbf{1}[y_{i,k}=c\land\hat y_{i,k}=c]}{\sum_{i\in M_{T1,k}}\mathbf{1}[y_{i,k}=c]}.
\end{equation}
The class denominator is the number of T1-matched reference events in class $c$. Thus a majority-class prediction cannot obtain a high tier score by ignoring rare labels.

Null labels are excluded only from the corresponding T2 or T3 denominator and never remove an event from the T1 denominator. T2 measures the associated state update; T3 measures the resulting task-relevant state at the environment's reference temporal granularity. Temporal matching remains anchored to the event timestamp $t_i$.

The benchmark tier labels instantiate these rules as follows. Go, Chess, and Gomoku use the quantized event-associated win-rate change for T2 and the post-event win-rate leader for T3. Doudizhu uses the hidden-hand PIMC event-associated win-rate change for T2 and the post-event landlord-versus-farmers leader for T3, with evaluator abstentions represented by null labels. CS2 uses the winner of the round containing the event for T2 and the cumulative round score after that round for T3, both carried on the event record rather than as separate per-event tier fields. DOTA~2 uses the change in parsed net-worth advantage over the aligned minute-level interval for T2 and the net-worth leader at its end for T3; these are encoded by the \texttt{minute\_gain} and \texttt{leader\_after} fields and use a $\pm1000$ gold dead band. StarCraft~II uses the banded event-associated change in resource-equivalent value for T2 and the banded resource-equivalent leader after resolution for T3. GFootball uses the goal-associated score-differential transition for T2 and the exact post-goal score pair for T3. Badminton uses the point-caused score-differential transition for T2 and the exact post-point score pair for T3.

Scores are first averaged equally over cases within each environment and then macro-averaged over the nine environments. Within a case, T2 and T3 denominators contain only valid T1-matched events for that tier. Infrastructure failures are excluded from model means, while valid zeroes and named protocol outcomes remain in the denominator.

\subsection{Human audit}
\label{app:human}

We conduct a small human audit to verify that the case videos contain enough evidence to reconstruct the target tuples. The participants were all adults aged 18 and older, with higher education backgrounds, and proficient in computer use. They are free to quit the process during any stages. Participants could freely pause, seek, and replay the full video and were given a working document for recording intermediate events and states. This setting is analogous to Protocol~B in allowing iterative inspection and persistent notes, and the final responses are evaluated with the same strict tuple-level F1. Due to limited resources, each participant evaluated 20 cases per environment only once. Participation was entirely voluntary, informed, and subject to unconditional withdrawal at any time. For Counter-Strike~2, we recruited two B+ players, approximately the top 25\% of the player population; the DOTA~2 and StarCraft~II audits used players ranked in the top 30\% of their respective ladders. For GFootball and Badminton, two non-specialist participants familiar with the basic rules served as the human baseline. The Go, Chess, Gomoku, and Doudizhu audits also used amateur participants because we could not recruit expert players for these four environments. Table~\ref{tab:human_audit} reports the resulting scores. Given the small and expertise-dependent participant pool, these results are a recoverability check rather than a population-level estimate or formal human ceiling.

Organizing the audit by the model bottlenecks identified in Figure~\ref{fig:tiers_micro} supports the three-way diagnosis. In the \emph{perception-bound} group, humans reach 98.6 on GFootball, 96.4 on Counter-Strike~2, and 96.2 on Badminton. Once their salient events are observed, the score-like state can be interpreted and tracked with little ambiguity; the much lower model scores therefore primarily reflect failure to recover the visual event stream. The \emph{reasoning-bound} group gives the cleanest separation: non-specialists obtain 100\% T1 reconstruction on Go, Chess, Gomoku, and Doudizhu, yet strict F1 falls to 51.6, 76.2, 63.2 and 56.5 because judging engine-derived win-rate changes and leaders requires game-specific expertise that our participants lacked. In the \emph{perception-and-reasoning-bound} group, experienced DOTA~2 and StarCraft~II players reconstruct nearly all T1 events, but small early- and mid-game resource advantages remain visually ambiguous, reducing strict F1 to 73.6 and 75.0. Human expertise thus removes most of the perceptual bottleneck in these two environments while leaving a residual state-inference bottleneck. Accordingly, the categories in Figure~\ref{fig:tiers_micro} describe the bottlenecks of current models rather than fixed intrinsic limits of the environments. Current models average below 10 points—significantly lower than the scores of amateur human players. Consequently, even for the vast majority of predictable events, models still exhibit severe deficiencies in perception, interpretation, and tracking.

\begin{table}[!htbp]
\centering
\small
\caption{Human audit under free video playback and note-taking, grouped by the model-bottleneck categories in Figure~\ref{fig:tiers_micro}. The score is the average of two people. T1 observations are reported qualitatively when exact tier scores were not retained. F1 is the strict tuple-level score on the $[0,100]$ scale.}
\label{tab:human_audit}
\adjustbox{max width=\linewidth}{%
\begin{tabular}{@{}lllr@{}}
\toprule
\textbf{Environment} & \textbf{Participant background} & \textbf{T1 reconstruction} & \textbf{F1$_{strict}$} \\
\midrule
\rowcolor{green!10}\multicolumn{4}{@{}l}{\emph{Perception-bound environments}} \\
GFootball        & Two non-specialists familiar with rules & Nearly complete              & 98.6 \\
Counter-Strike~2 & Two B+ players (approximately top 25\%) & Nearly complete & 96.4 \\
Badminton        & Two non-specialists familiar with rules & Nearly complete              & 96.2 \\
\rowcolor{red!8}\multicolumn{4}{@{}l}{\emph{Reasoning-bound environments}} \\
Go               & Two amateur players                         & 100\%           & 51.6 \\
Chess            & Two amateur players                           & 100\%           & 76.2 \\
Gomoku           & Two amateur players                            & 100\%           & 63.2 \\
Doudizhu         & Two amateur players                            & 100\%           & 56.5 \\
\rowcolor{gray!15}\multicolumn{4}{@{}l}{\emph{Perception-and-reasoning-bound environments}} \\
DOTA~2           & Two ladder players (top 30\%)               & Nearly complete & 73.6 \\
StarCraft~II     & Two ladder players (top 30\%)               & Nearly complete & 75.0 \\
\bottomrule
\end{tabular}%
}
\end{table}

\subsection{Per-environment uncertainty and outcome accounting}
\label{app:uncertainty}

Table~\ref{tab:ci} reports how sensitive each environment-level result is to the particular 20 cases selected for that environment. For each of $10^{4}$ bootstrap replicates, we sample case identifiers with replacement within an environment. All available system results for a sampled case move together, preserving the fact that the same case may be easy or difficult for several systems. We recompute each system's mean on the sampled cases and then average the system means with equal weight, matching the aggregation used in the main results. A system contributes only cases with archived scores; missing cells are not imputed. The 2.5th and 97.5th percentiles form the 95\% confidence interval, and a fixed seed makes the calculation reproducible.

These intervals answer one question: how much would the reported environment mean change if a different set of cases had been selected? They do not measure run-to-run variation of a model or harness. Each system--case cell has one effective run, so randomness in an agent's tool-use trajectory or in a video-only response is outside the interval. The intervals should therefore be interpreted as case-sampling uncertainty, not as repeatability bounds.

The paired column measures the effect of changing the protocol while holding both the system and case fixed. We first compute agentic minus video F1 for every system--case pair scored under both protocols, and then bootstrap those matched cases. Unmatched cells are excluded from the paired analysis. Because the paired estimate uses this common subset, it need not equal the difference between the two separately aggregated protocol means.

\begin{table}[!htbp]
\centering
\caption{Environment-level strict F1 with 95\% case-bootstrap confidence intervals and paired agentic-minus-video differences. Each mean gives equal weight to systems; $10^{4}$ resamples and a fixed seed are used. Protocol A includes eight to ten systems depending on archived coverage, Protocol B includes all ten, and the paired estimate uses only matched system--case cells. The intervals quantify case-sampling uncertainty, not run-to-run variation.}
\label{tab:ci}
\adjustbox{max width=\linewidth}{%
\begin{tabular}{@{}l r l r l r l@{}}
\toprule
& \multicolumn{2}{c}{\textbf{Video}} & \multicolumn{2}{c}{\textbf{Agentic}} & \multicolumn{2}{c}{\textbf{Paired} $\Delta$} \\
\cmidrule(lr){2-3}\cmidrule(lr){4-5}\cmidrule(lr){6-7}
\textbf{Environment} & \textbf{F1} & \textbf{95\% CI} & \textbf{F1} & \textbf{95\% CI} & $\Delta$ & \textbf{95\% CI} \\
\midrule
Go        &  2.4 & [1.8, 2.9]   & 28.2 & [24.6, 30.8] & $+24.9$ & [$+21.8$, $+28.1$] \\
Chess     & 12.3 & [8.7, 15.7]  & 54.8 & [50.0, 60.0] & $+43.5$ & [$+38.1$, $+47.7$] \\
Gomoku    &  7.0 & [4.4, 9.1]   & 22.4 & [19.4, 25.8] & $+16.1$ & [$+13.4$, $+17.9$] \\
Doudizhu  & 13.3 & [10.7, 15.8] & 18.8 & [14.7, 20.4] & $+4.4$  & [$+2.0$, $+7.5$]   \\
CS2       &  3.9 & [2.6, 6.0]   & 22.8 & [17.7, 25.0] & $+18.4$ & [$+16.0$, $+23.3$] \\
DOTA~2    &  0.4 & [0.3, 1.2]   &  7.7 & [5.9, 9.4]   & $+5.9$  & [$+4.5$, $+7.3$]   \\
SC2       &  0.7 & [0.2, 1.2]   &  3.7 & [1.9, 4.1]   & $+2.9$  & [$+2.0$, $+4.0$]   \\
GFootball &  0.4 & [0.1, 0.7]   & 18.7 & [14.8, 21.6] & $+21.5$ & [$+18.1$, $+25.1$] \\
Badminton &  0.6 & [0.4, 1.0]   & 11.1 & [9.2, 13.1]  & $+13.0$ & [$+10.6$, $+15.7$] \\
\bottomrule
\end{tabular}%
}
\end{table}

The table shows three clear patterns. First, under the visual protocol, DOTA~2, StarCraft~II, GFootball, and Badminton remain near zero, and their confidence intervals stay below those of the board environments. Their low scores are therefore not explained by a few unusually difficult sampled cases. Second, the agentic protocol improves performance on every environment: all paired confidence intervals are above zero. The gains are largest for Chess ($+43.5$) and GFootball ($+21.5$), but small for Doudizhu ($+4.4$) and StarCraft~II ($+2.9$). These two small gains have different meanings: Doudizhu begins with the strongest video score and changes little, whereas StarCraft~II remains near the floor under both protocols. Third, Chess is the only environment whose agentic mean exceeds 50, and its relatively wide interval indicates substantial variation in case difficulty.

We distinguish infrastructure failures from valid model outcomes when computing these means. Infrastructure failures, such as routing, API, or sandbox errors, are rerun and excluded from the reported averages. Budget exhaustion and format or protocol errors remain valid model outcomes because the system failed to produce a compliant answer within the stated constraints. Runs that fail the egress audit are voided, assigned zero, and retained in the denominator; this classification takes priority over all others. Among all more than 4000 archived runs, 88.28\% completed cleanly, 3.27\% were infrastructure failures, 6.71\% were budget-bound, 1.68\% ended in a format or protocol outcome, and 0.06\% were voided. 

All reported agentic scores come from the final hardened sandbox in Appendix~\ref{app:integrity}. The initial file-whitelist check passed for all 1800 archived agentic runs, so none began with an unexpected file. Two runs attempted package installation, failed the egress audit, and were voided. Twenty runs from one system--environment cell never committed an answer; they are recorded as \texttt{never\_wrote} and scored zero. Another 28 runs created files whose names triggered the banned-artifact check. The audit confirmed that these were model-created working files, such as board transcriptions, rather than hidden records seeded into the sandbox. These checks separate genuine model failures from infrastructure problems and from possible leakage.

\section{Sandbox and protocol integrity}
\label{app:integrity}

The two protocols share the same task definitions, reference data, judge, event vocabulary, and temporal tolerance. Protocol A receives only a globally fixed sampled frame sequence with explicit timestamps and a single structured-output request. It permits no tools, code execution, external media inspection, replay, sidecar, ground truth, or second model call. Protocol B receives the full rendered video, neutral media tools, the instruction, a pre-seeded answer file, and the native tools of the evaluated harness. It permits multi-turn analysis but not game interaction, replay, timing sidecars, value curves, ground truth, judge or self-check scripts, engine weights, external model downloads, or any channel that can retrieve the source contest.

\subsection{Sandbox construction and permissions}

Each Protocol B run is placed in an opaque directory outside the repository. The video is copied to the neutral name \texttt{match.mp4}; the only other seeded files are \texttt{ffmpeg.exe}, \texttt{ffprobe.exe}, \texttt{instruction.md}, and the harness-prepared \texttt{answer.json}. Claude Code additionally receives an isolated configuration directory. The harness checks exact set equality between the top-level directory and this whitelist before spending tokens. The answer seed must have the same root key as the judge, otherwise the run fails closed.

The child process starts with the sandbox as its working directory and first entry on \texttt{PATH}. Inherited Claude Code state, process identifiers, API keys, and context-related environment variables are scrubbed before the harness applies the pinned model and context configuration. The agent is instructed to write and repeatedly overwrite \texttt{answer.json}; text in the final chat response is not an answer. The sandbox does not contain \texttt{.env}, \texttt{probe.py}, ground-truth files, judge code, replay files, engine weights, or timing sidecars.

After the run, an independent banned-artifact walk checks all agent-produced files outside the harness configuration directory. It rejects ground-truth names, replay extensions, evaluator or judge artifacts, manifests, traces, and other answer-bearing side channels. The exact whitelist audit and the post-run walk are separate checks: the first catches incorrect seeding, while the second catches artifacts created during execution.

\subsection{Budgets, reproducibility, and failure classes}

The reported agentic condition allows at most 200 interaction turns and 2 hours per case. The budget is a wall-clock limit, and reaching either limit is recorded as a valid budget-bound result rather than silently retried. Prompts are frozen, temperature is fixed at the lowest supported setting, and each $(\text{system},\text{case})$ pair has one effective run. A rerun is permitted only for a documented infrastructure failure and receives a new result record.

Ground truth is never mounted into the prompt or sandbox; scoring is performed offline by the frozen judge. Each run records the model, harness, deployment, budget source, sampling, stop reason, usage, answer-file state, sandbox audit results, egress audit, code version, judge hash, instruction hash, and answer-seed hash. Aggregation distinguishes infrastructure failures such as routing, proxy, API, sandbox, or harness errors from valid outcomes such as \texttt{never\_wrote}, \texttt{budget\_bound}, \texttt{format\_error}, and ordinary capability failures. A failed egress audit has priority over all other classifications and voids the run.

These controls are intended to make the benchmark measure video understanding rather than access to hidden records, an evaluator, a second model, or an accidentally different prompt. They also make a zero inspectable: the result is either a valid low score or a named protocol or infrastructure failure with evidence preserved for audit.

\section{Game-specific construction details}
\label{app:game_details}

This appendix documents the source-to-video adapters for the nine environments. Each subsection gives the source record or simulator, the state used for ground truth, the three tiers, the rendered input, and an illustrative six-frame strip. Each strip is decoded from a single case video at six increasing timestamps spanning that case, so the panels are excerpts of an evaluation input rather than additional evaluation cases. The Protocol~A input is a 1-FPS-priority sequence of individual still frames, uniformly spaced from the start of the video and capped at 256 frames spanning the full duration once the case is longer than 256 seconds; Protocol~B receives the full rendered video in the opaque sandbox. Each listing is the first tier-bearing ground-truth record of that same illustrated case, reproduced verbatim except that fields unrelated to the three tiers are elided, so the field names and label alphabets are the actual ones rather than a stylized schema.

\lstset{basicstyle=\ttfamily\scriptsize,columns=fullflexible,breaklines=true,frame=single,backgroundcolor=\color{gray!8},showstringspaces=false,keepspaces=true}

\newcommand{\appdfigure}[4]{%
\begin{center}
#2
\captionsetup{type=figure,hypcap=false}
\captionof{figure}{\textbf{#3}, case #4. All six frames are decoded from that one case video at the timestamps printed on each panel, read left to right and then top to bottom. The panels are the rendered visual channel as a model receives it; no ground truth, value trace, or judge output is overlaid. Elements that the render deliberately retains are stated per environment in the text above.}
\label{fig:appd-#1}
\end{center}
\smallskip}

\subsection{Go}
\label{appd:go}
Go cases use SGF records exported from the OGS public API (fixture game 27400000) and a deterministic board-only renderer. The adapter replays each move, records the move or capture event, and evaluates every prefix with the pinned KataGo configuration~\citep{wu2019katago}. The rendered source is a plain $19\times19$ board; coordinates, move numbers, names, clocks, captures, komi, and evaluation bars are removed. T1 identifies the move or capture, its time, colour, and board entities. T2 is the quantized signed Black win-rate change associated with the move. T3 is the win-rate leader after the move. The source service is \url{https://online-go.com} (accessed: 2026-09-04); the benchmark fixture is identified by OGS game ID 27400000.

\appdfigure{go}{\includegraphics[width=0.78\linewidth]{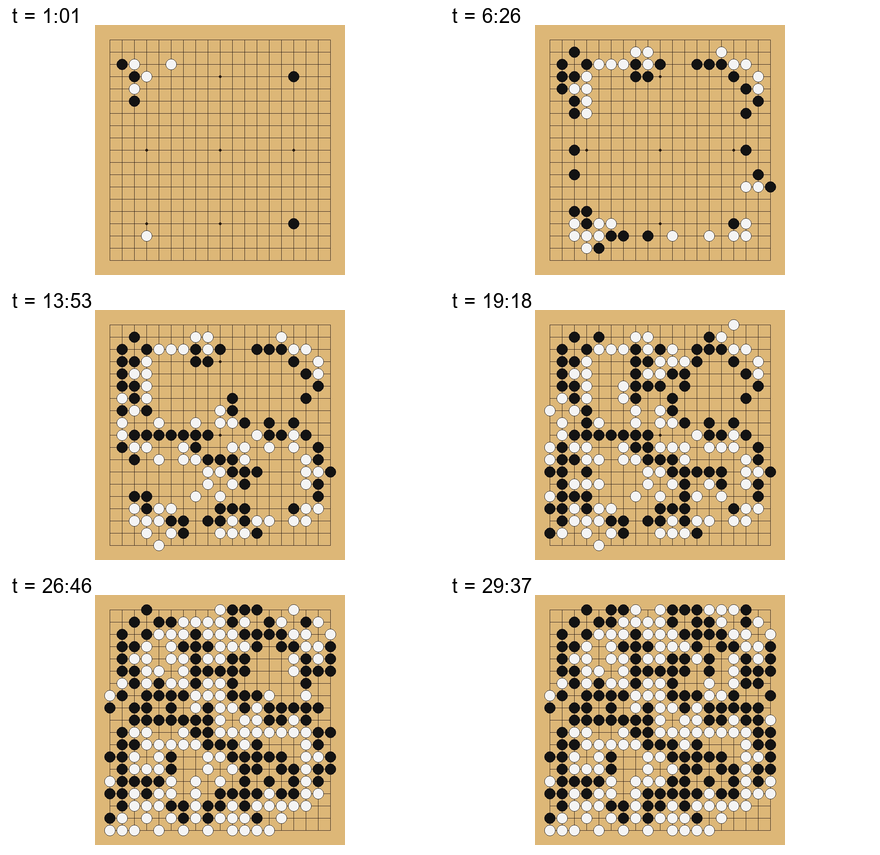}}{Go}{OGS~27400000}
\begin{lstlisting}
{
  "events": [
    {"type": "move", "idx": 1, "color": "B", "coord": "Q4",
     "game_clock_s": 2.0, "wr_swing": "equal", "leader_after": "equal"}
  ]
}
\end{lstlisting}

\subsection{Chess}
\label{appd:chess}
Chess cases use the public Lichess export for fixture cZWtZEBE and are replayed with per-ply timing on a plain $8\times8$ board. Stockfish supplies the pinned White-oriented expected-score/WDL value trajectory~\citep{stockfish}; draws map to the neutral label. The renderer removes coordinates, move numbers, player names, clocks, result text, captured-piece trays, and evaluation bars. T1 identifies the move or capture, its time, moving colour, and destination square; the benchmark contract records the destination only, not the origin square. T2 is the quantized change in the White-oriented value associated with the ply. T3 is the post-ply leader. The source service is \url{https://lichess.org} (accessed: 2026-09-04); the benchmark fixture is identified by game ID cZWtZEBE and its public game export.

\appdfigure{chess}{\includegraphics[width=0.78\linewidth]{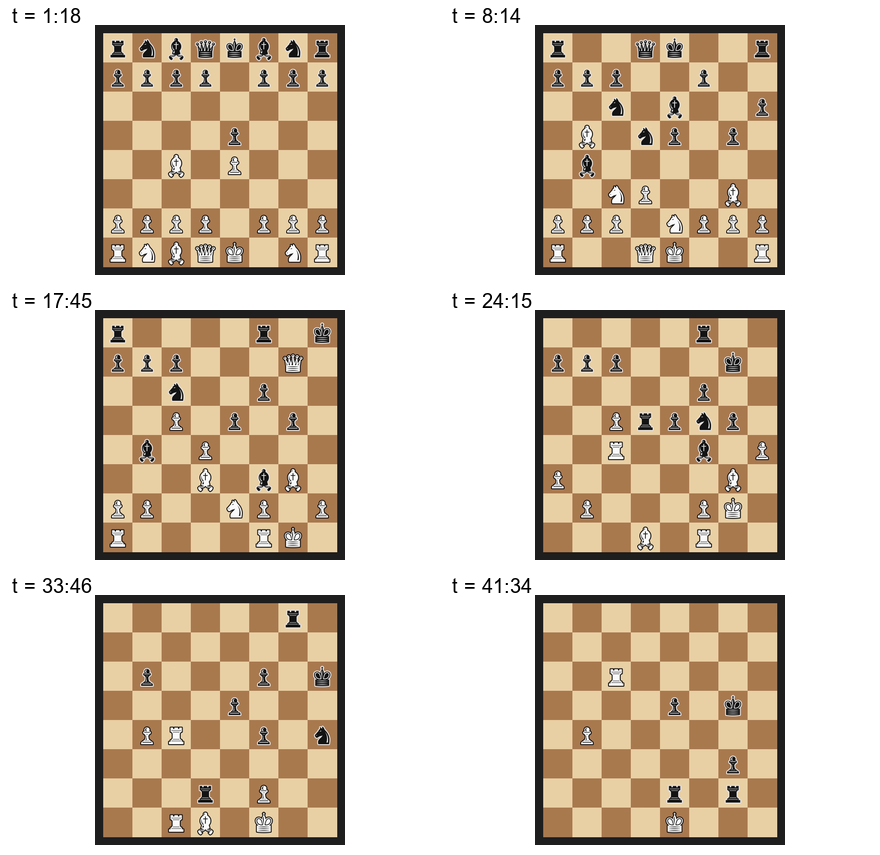}}{Chess}{Lichess~\texttt{cZWtZEBE}}
\begin{lstlisting}
{
  "events": [
    {"type": "move", "idx": 1, "color": "W", "coord": "E4",
     "game_clock_s": 2.0, "wr_swing": "equal", "leader_after": "equal"}
  ]
}
\end{lstlisting}

\subsection{Gomoku}
\label{appd:gomoku}
Gomoku cases use competition PSQ records and are replayed on a plain $15\times15$ board. Rapfi is the pinned evaluator~\citep{rapfi}, not the source-record provider. The adapter converts the raw side-to-move trajectory to the benchmark's tempo-neutral trajectory, preventing alternating move parity from becoming a label shortcut. T1 identifies the placed stone and its coordinate. T2 is the quantized local win-rate change associated with the move. T3 is the post-move leader. The evaluator repository is \url{https://github.com/dhbloo/rapfi} (accessed: 2026-09-04); the source records are competition PSQ files. The renderer removes coordinates, move numbers, names, clocks, and engine annotations.

\appdfigure{gomoku}{\includegraphics[width=0.78\linewidth]{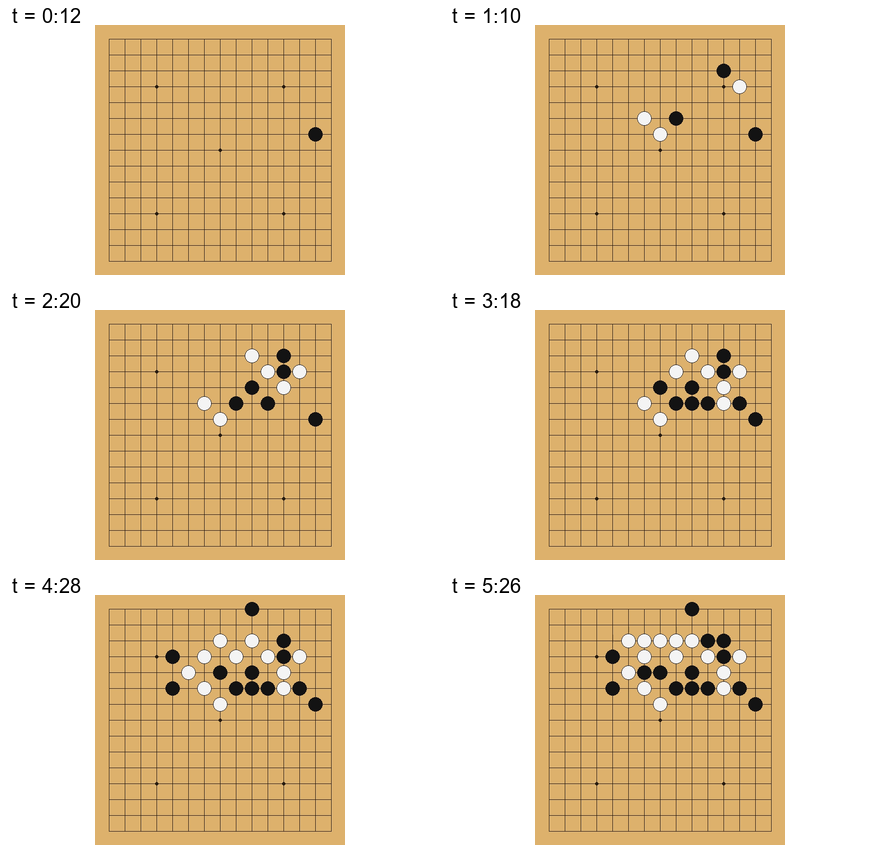}}{Gomoku}{PSQ record~\texttt{0\_11\_6\_2}}
\begin{lstlisting}
{
  "events": [
    {"type": "move", "idx": 1, "color": "B", "coord": "N7",
     "game_clock_s": 2.0, "wr_swing": "W", "leader_after": "W"}
  ]
}
\end{lstlisting}

\subsection{Doudizhu}
\label{appd:doudizhu}
Doudizhu cases are generated from DouZero-WP greedy self-play deals (\texttt{tau=0}) using the released DouZero environment and weights~\citep{douzero}, then rendered as a tabletop game. The landlord hand and bottom cards are face up; farmer hands and ranks remain hidden, with opaque card backs revealing only their counts. The adapter emits every non-pass play with player, combination type, and card count. Hidden-hand determinization followed by adaptive 400--4000-world sampling and greedy rollouts with the released DouZero-WP policy provides the value trajectory~\citep{long2010pimc}; the PIMC estimator is our benchmark implementation, not a DouZero component. T1 identifies the non-pass play, player, combination type, and card count. T2 is the quantized win-rate change associated with the play, with landlord, farmers, and neutral labels. T3 is the landlord, farmers, or equal leader after the play. Abstained tier labels are stored as \texttt{null} and excluded from the corresponding T2 or T3 denominator and strict comparison, while the row remains a T1 event that the model must answer. Sources: \citep{douzero,douzero_weights}; \url{https://github.com/kwai/DouZero} and \url{https://huggingface.co/palemoky/douzero-baselines} (accessed: 2026-09-04). The renderer removes numeric scores, timers, move counters, probability displays, and farmer-hand ranks.

\appdfigure{doudizhu}{\includegraphics[width=0.78\linewidth]{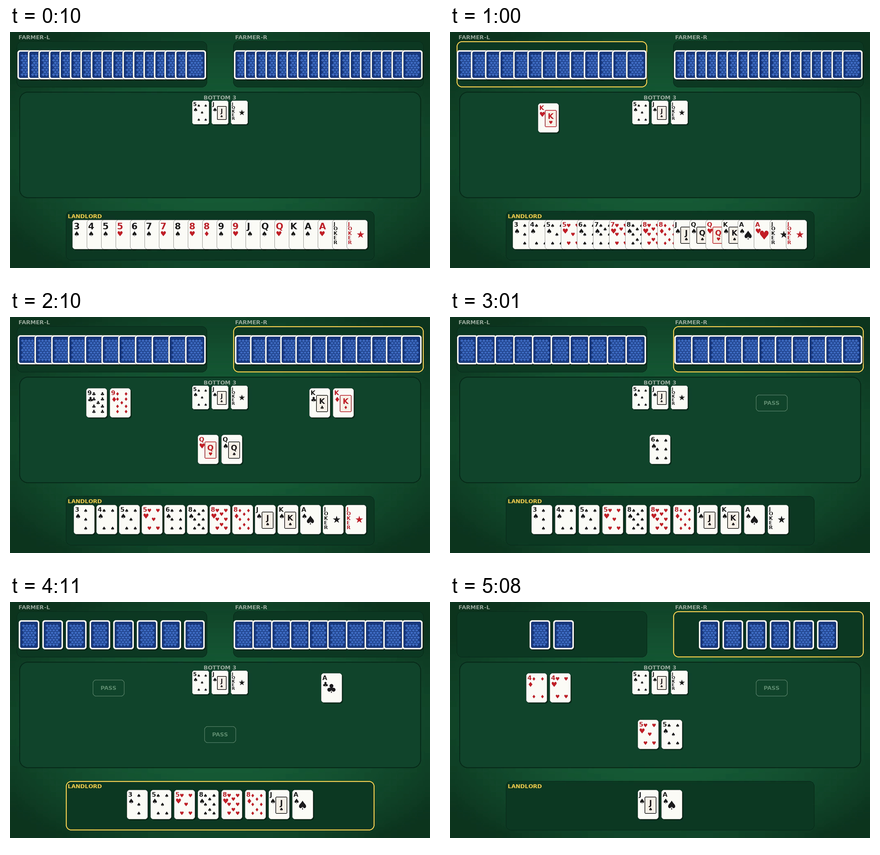}}{Doudizhu}{\texttt{dzh-01}}
\begin{lstlisting}
{
  "events": [
    {"idx": 1, "game_clock_s": 15.0, "player": "landlord", "move_type": "pair",
     "num_cards": 2, "impact": "neutral", "leader_after": "farmers"}
  ]
}
\end{lstlisting}

\subsection{Counter-Strike 2}
\label{appd:cs2}
Counter-Strike~2 cases begin with public professional-match demo (\texttt{.dem}) files published on HLTV, parsed at their native tick rate by our own parser and re-rendered as clean top-down radar views. The adapter aligns kills, eliminations, plants, defuses, explosions, and saves with the round state. T1 identifies a type-specific event---\texttt{kill}, \texttt{bomb\_planted}, \texttt{bomb\_defused}, \texttt{bomb\_exploded}, \texttt{elimination}, or \texttt{save}---with its time and required side, player, or objective attributes. On a \texttt{kill} the side and the dot number~1--10 identify the victim rather than the killer, because the death marker is deliberately rendered without team colour or number. T2 is the winner of the round containing the event, and T3 is the cumulative score after that round; the CS2 contract carries both on the event record rather than exposing separate per-event fields. The render preserves map geometry, player motion, bomb state, and event-relevant effects, but removes the HUD, scoreboard, round counter, kill feed, economy, names, and audio; the retained bomb/countdown clock is the only textual timing aid. The game itself is \url{https://www.counter-strike.net} (accessed: 2026-09-04). Cases are identified by their match slugs; the tournament and stage behind each demo are not recorded in the benchmark.

\appdfigure{cs2}{\includegraphics[width=0.78\linewidth]{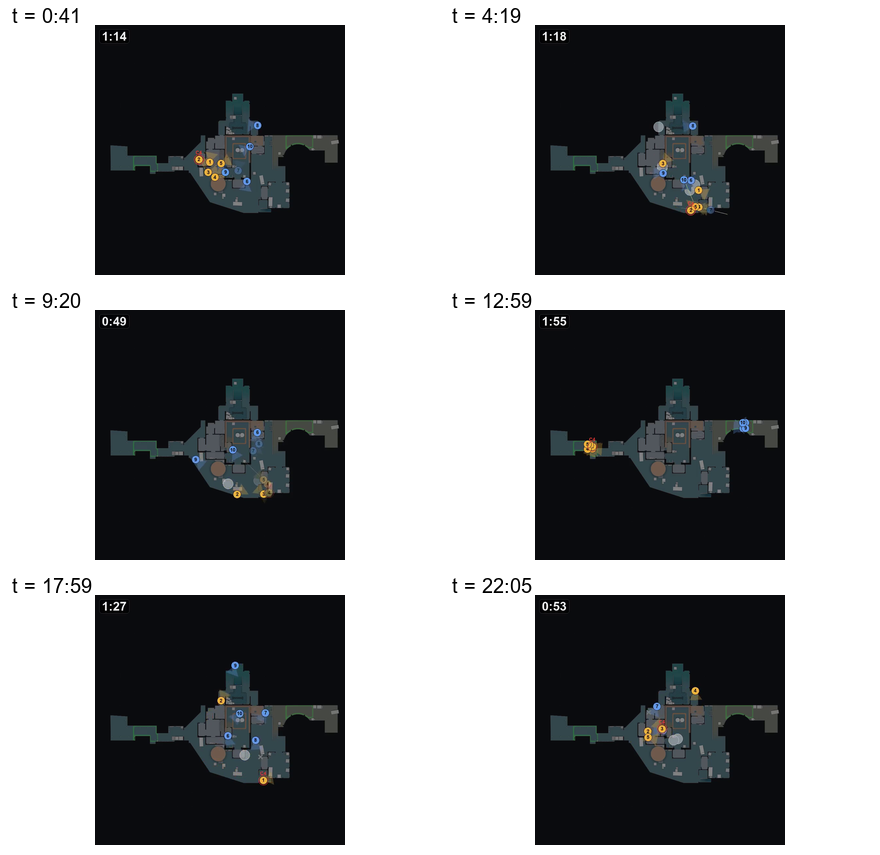}}{Counter-Strike 2}{\texttt{9z-vs-legacy-m2-nuke}}
\begin{lstlisting}
{
  "events": [
    {"round_no": 1, "type": "kill", "video_ts": 49.812, "side": "CT",
     "number": 6, "round_winner": "CT", "score_after": {"A": 0, "B": 1}}
  ]
}
\end{lstlisting}

\subsection{DOTA 2}
\label{appd:dota2}
DOTA~2 cases use Valve replays and parsed match records from the OpenDota ecosystem. The adapter extracts six event types: first blood, teamfights, towers, barracks, Roshan, and the ancient, with type-specific side, killer or winner, death-count, tier, lane, and objective attributes, and aligns them to the real-time minimap render. The task-relevant state is the parsed net-worth advantage, not a learned win-probability model. T1 identifies the event and its required attributes. T2 quantizes the change in net-worth advantage over the aligned minute-level interval associated with the event; T3 identifies the net-worth leader at that interval's end. Both labels are derived from the minute-level gold curve, use the fixed $\pm1000$ gold dead band, and are stored in the fields \texttt{minute\_gain} and \texttt{leader\_after}, respectively. The render uses the real terrain minimap and game sprites but removes the HUD, scoreboard, clock, net-worth graph, names, and objective counters. Sources: \url{https://www.opendota.com} and \url{https://github.com/odota} (accessed: 2026-09-04).

\appdfigure{dota2}{\includegraphics[width=0.78\linewidth]{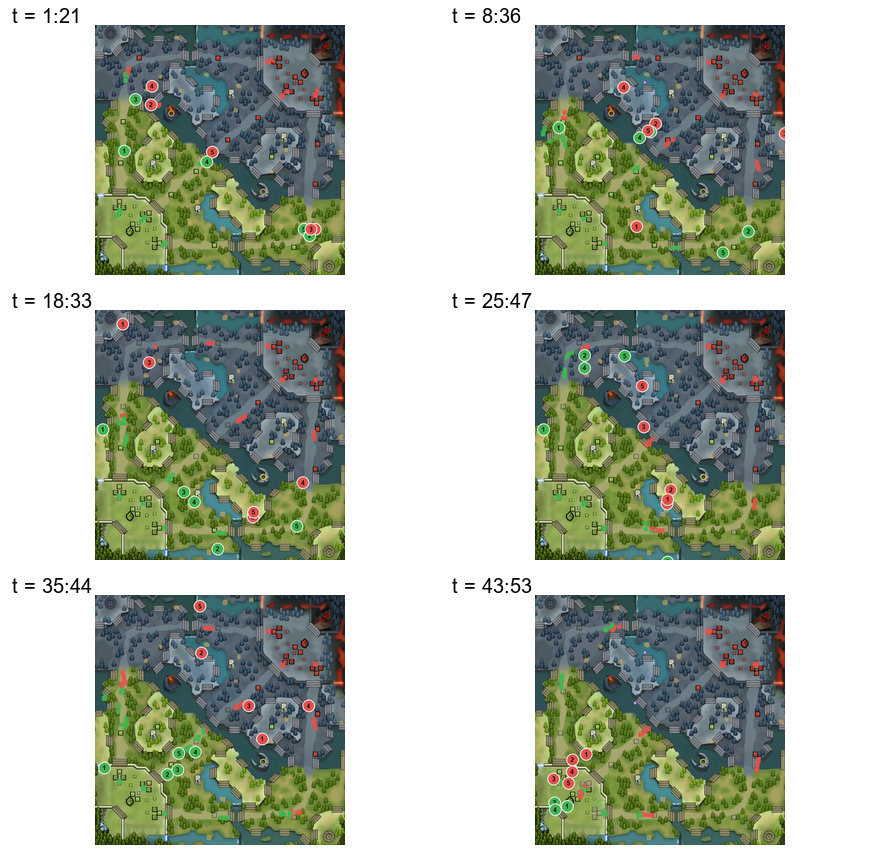}}{DOTA 2}{\texttt{dota-01}}
\begin{lstlisting}
{
  "events": [
    {"type": "firstblood", "time_s": 9, "video_ts": 17.0,
     "killer_side": "dire", "minute_gain": "even", "leader_after": "even"}
  ]
}
\end{lstlisting}

\subsection{StarCraft II}
\label{appd:sc2}
StarCraft~II cases are generated from engine-controlled replay files and rendered as full-map isometric observer views rather than literal vertical top-down views. The replay tracker stream supplies battle and building-destruction events together with resource-equivalent value records. T1 identifies the target event, its time, and the coarse map region in which it occurs: the playable area is quantized into a $3\times3$ world grid whose cells are written \texttt{r\{row\}c\{col\}}, so the label names the region a viewer would point at rather than a pixel position. T2 is the banded change in resource-equivalent value associated with the event (\texttt{fight\_result}); T3 is the banded resource-equivalent leader after resolution (\texttt{leader\_after}). The value combines banked minerals and gas, live army value, economy value, and technology value. The renderer removes resource counters, production panels, minimaps, names, and direct state readouts. The replay protocol source is \url{https://github.com/Blizzard/s2client-proto} (accessed: 2026-09-04). Each case contains one $1536\times1536$ video at 5.6~FPS covering the whole map from a fixed viewpoint, together with its ground-truth file; the $3\times3$ tiling that the render pipeline uses internally is not part of the benchmark media, and the video is not a normal spectator HUD capture.

\appdfigure{sc2}{\includegraphics[width=0.78\linewidth]{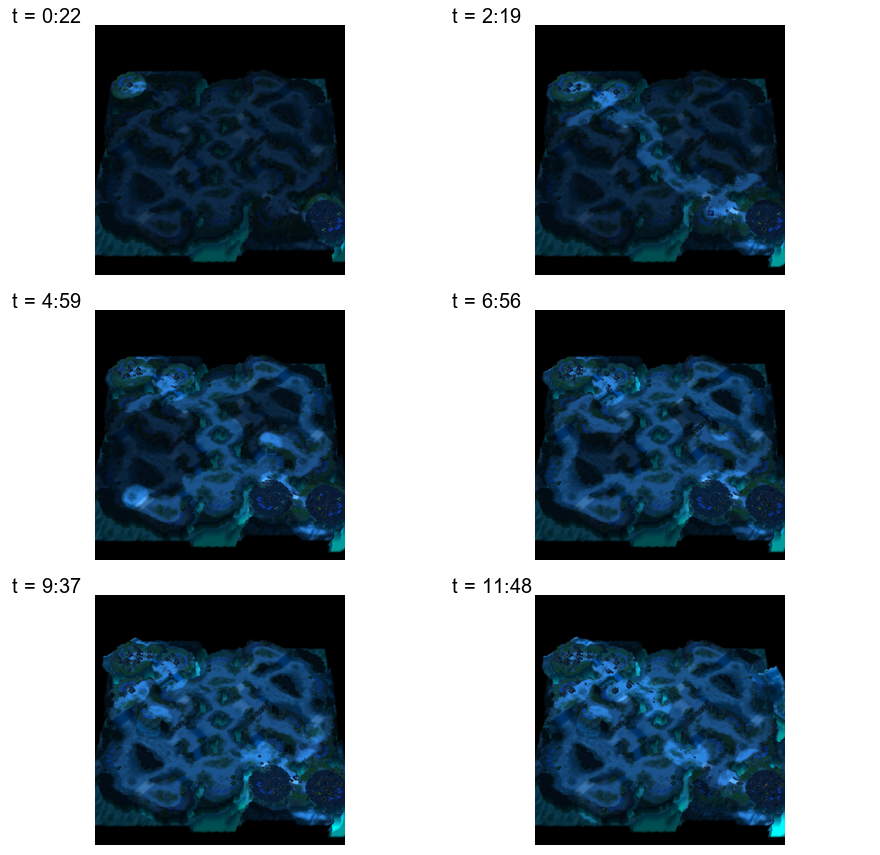}}{StarCraft II}{\texttt{g00}}
\begin{lstlisting}
{
  "events": [
    {"game_clock_s": 235, "type": "battle", "entity": "r0c0",
     "fight_result": "B", "leader_after": "equal"}
  ]
}
\end{lstlisting}

\subsection{GFootball}
\label{appd:gfootball}
GFootball cases are generated from randomized Google Research Football simulator episodes~\citep{gfootball}. The adapter reads score transitions and the engine goal log, including the scoring side and scorer, then renders the synchronized full $1280\times720$, 24-FPS default inclined 3D broadcast view in real time. T1 identifies the goal, its time, scoring side as \texttt{L} or \texttt{R}, scorer surname, and own-goal flag. T2 is the four-class score-differential change associated with the goal, \texttt{taken}, \texttt{extended}, \texttt{levelled}, or \texttt{narrowed}, determined from the pre-goal score difference. T3 is the exact native score pair after the goal, so it is scored by exact match rather than by a three-way leader label. The production render suppresses the scoreboard, match clock, and goal subtitles; the minimap and the floating scorer-name label remain visible because they are part of the benchmark visual contract. Development previews may show other HUD elements and are not evaluation inputs. Source: \url{https://github.com/google-research/football} (accessed: 2026-09-04).

\appdfigure{gfootball}{\includegraphics[width=0.78\linewidth]{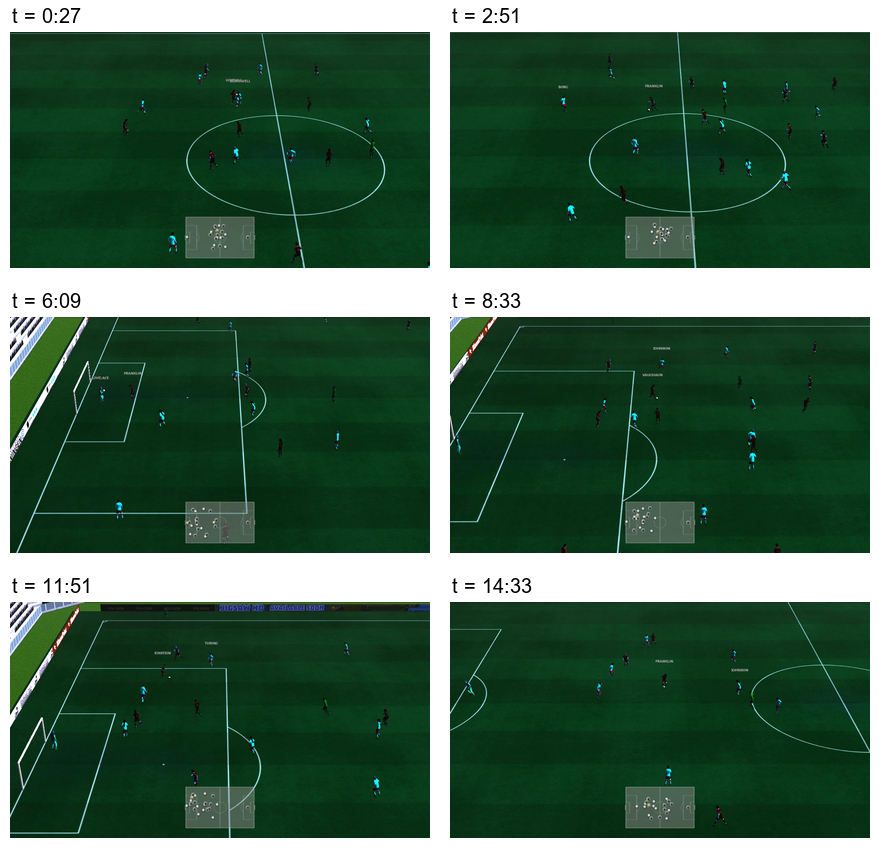}}{GFootball}{\texttt{gf-01}}
\begin{lstlisting}
{
  "events": [
    {"type": "goal", "idx": 1, "game_clock_s": 78.458, "scoring_side": "L",
     "scorer_name": "Blackwell", "impact": "taken", "score_after": [1, 0],
     "own_goal": false}
  ]
}
\end{lstlisting}

\subsection{Badminton}
\label{appd:badminton}
Badminton cases use ShuttleSet stroke records~\citep{shuttleset} and a fixed-camera synthetic 3D renderer built from real-match trajectories. CoachAI-Projects is cited as the upstream data and environment source~\citep{coachai}, not as our renderer. The adapter groups strokes into points, records the scoring side and terminal landing-zone event, and carries the native score through the rally. T1 identifies the point, its time, scoring side as \texttt{near} or \texttt{far} relative to the fixed camera, and one of \texttt{net}, \texttt{out}, \texttt{in\_front}, \texttt{in\_mid}, or \texttt{in\_back}. T2 is the four-class score-differential change associated with the point; T3 is the exact score pair after the point. The fixed renderer is $1280\times720$ at 15 FPS, preserves player, racket, and shuttle motion, and removes broadcast scoreboards, names, clocks, and annotations; it is a synthetic render, not a source replay. Sources: \url{https://github.com/wywyWang/CoachAI-Projects} and \url{https://arxiv.org/abs/2306.04948} (accessed: 2026-09-04).

\appdfigure{badminton}{\includegraphics[width=0.78\linewidth]{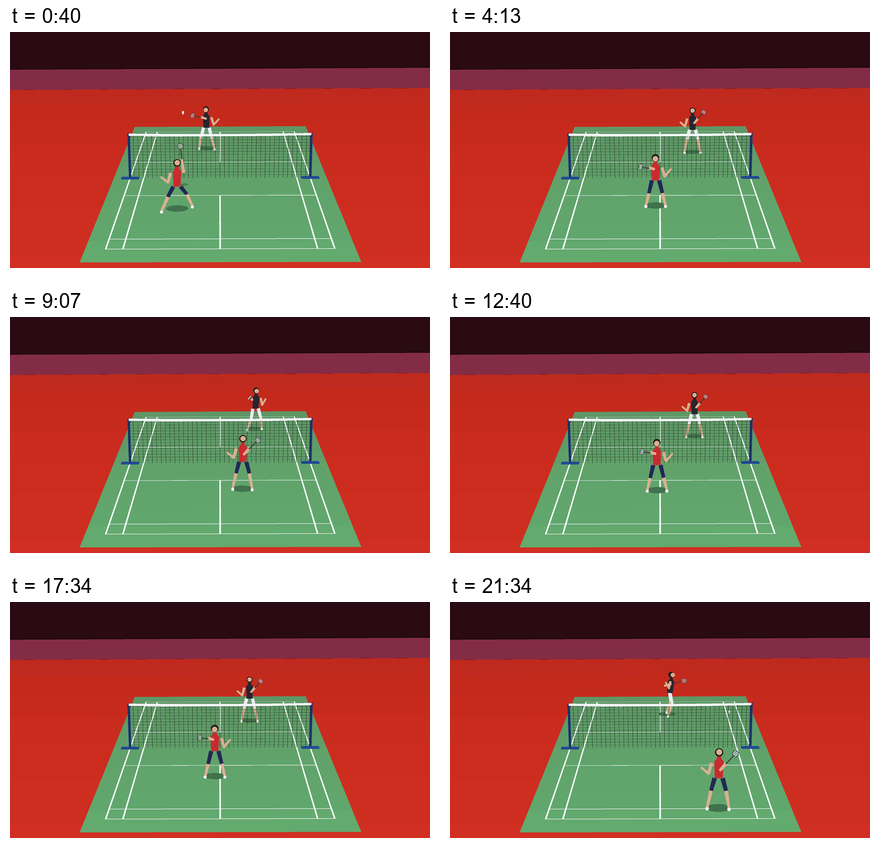}}{Badminton}{\texttt{badminton-01}}
\begin{lstlisting}
{
  "events": [
    {"type": "point", "idx": 1, "game_clock_s": 5.84, "scoring_side": "near",
     "zone": "out", "impact": "taken", "score_after": [1, 0]}
  ]
}
\end{lstlisting}

Across all nine adapters, source records and simulator traces are used only during construction. Protocol~A exposes only the rendered video and its timestamps, while Protocol~B additionally provides the rendered video, instructions, and a pre-seeded answer skeleton; neither protocol exposes the source replay, value trace, ground truth, or judge. Resolution is reported per environment as a property of the rendered media. In the Protocol~A transport, sampled frames preserve aspect ratio, are limited to 768 pixels in width, and are never upscaled when the source is smaller.

\end{document}